\PassOptionsToPackage{dvipsnames,table}{xcolor}
\documentclass{saladteam}

\usepackage{bbm}
\usepackage{booktabs,multirow,array}
\usepackage{nicefrac}
\usepackage{wrapfig}
\usepackage{needspace}
\newsavebox{\schemaconceptfigurebox}
\usepackage{pifont}
\usepackage{fancyvrb,fvextra}
\usepackage{siunitx}
\usepackage{url}
\usepackage{hyperref}
\hypersetup{
  colorlinks=true,
  linkcolor=SaladLink,
  citecolor=SaladLink,
  urlcolor=SaladLink,
  pdftitle={Tracing the Cascade: A Topology-Aware Evaluation Framework for Scientific Agent Hallucinations},
  pdfauthor={Xinshun Feng, Ziqi Miao, Lijun Li, Jing Shao}
}

\definecolor{api}{HTML}{E9C46A}
\definecolor{citeblue}{HTML}{4A7C9B}
\newcommand{\cmark}{\textcolor{ForestGreen}{\ding{51}}}
\newcommand{\xmark}{\textcolor{BrickRed}{\ding{55}}}
\fvset{breaklines=true,breakanywhere=true,fontsize=\small}

\newcommand{\answerYes}[1][]{\textcolor{blue}{[Yes]#1}}
\newcommand{\answerNo}[1][]{\textcolor{orange}{[No]#1}}
\newcommand{\answerNA}[1][]{\textcolor{gray}{[N/A]#1}}
\newcommand{\answerTODO}[1][]{\textcolor{red}{\bfseries [TODO]}}
\newcommand{\justificationTODO}[1][]{\textcolor{red}{\bfseries [TODO]}}

\saladrunningtitle{SCHEMA: Tracing the Cascade: A Topology-Aware Evaluation Framework for Scientific Agent Hallucinations}

\title{Tracing the Cascade: A Topology-Aware Evaluation Framework for Scientific Agent Hallucinations}

\author{%
  Xinshun Feng\textsuperscript{1,2,*}\thanks{Work done during an internship at Shanghai Artificial Intelligence Laboratory.} \quad
  Ziqi Miao\textsuperscript{2,*} \quad
  Lijun Li\textsuperscript{2,*,\textdagger} \quad
  Jing Shao\textsuperscript{2,\textdagger} \\
  \textsuperscript{1}Beihang University, Beijing, China\\
  \textsuperscript{2}Shanghai Artificial Intelligence Laboratory, Shanghai, China\\
  {\normalfont\texttt{lilijun@pjlab.org.cn}}\\
  {\normalfont\small
    \textsuperscript{*}Equal contribution.
    \quad
    \textsuperscript{\textdagger}Corresponding authors.}
}

\begin{document}

\setcounter{footnote}{2}
\maketitle

\begin{abstract}
Large language model (LLM) agents are increasingly deployed in scientific research, where reliability is critical, and the underlying knowledge is densely interconnected.
In such settings, hallucinations are particularly damaging: a single erroneous claim on a foundational concept can propagate through multi-step reasoning and corrupt entire trajectories.
Existing hallucination benchmarks largely operate at the surface level, treating facts in isolation and relying on uniform accuracy metrics that ignore this topological structure.
We address this gap with \textsc{Schema}, the first evidence-grounded, topology-aware evaluation framework for hallucinations in scientific agents.
\textsc{Schema} automatically constructs scientific concept graphs from benchmark seeds and literature evidence, synthesizes graph-grounded tasks spanning claim verification, multi-hop reasoning, open-ended explanation, and experimental code generation, and evaluates agents with two complementary diagnostics.
A trajectory hallucination pipeline audits intermediate reasoning at scale via a topology-weighted severity score ($\mathrm{HS}^w$), while a multi-agent counterfactual attribution module pinpoints the causal mechanism behind selected failures.
\textsc{Schema} reveals that hallucinations concentrate at a small set of highly connected knowledge hubs, and that final-answer accuracy decouples from trajectory honesty—models often reach correct conclusions through structurally flawed reasoning.
These results indicate that for high-stakes scientific applications, terminal accuracy alone is an insufficient signal of agent reliability, motivating mechanism-level evaluation grounded in knowledge topology\footnote{Code is available at \url{https://github.com/circles-post/SCHEMA}}.
\end{abstract}

\section{Introduction}

Large language model (LLM) agents have rapidly emerged as a powerful paradigm for open-ended problem-solving~\citep{wang2024survey, xi2025rise, yang2026toward}.
These agents are increasingly deployed in knowledge-intensive domains, particularly scientific research~\citep{lu2024ai, xu2025probing}.
However, hallucination remains a critical bottleneck as agents orchestrate complex scientific workflows~\citep{huang2025survey}: errors can originate from incorrect answers, unproductive tool invocations, or fabricated evidence, and a single such error propagates through downstream reasoning, fatally undermining system trustworthiness~\citep{du2024improving, zhu2026deep, robertson2026kghalubench}.

Recent efforts in hallucination detection have shifted from standalone LLMs to autonomous agents, with~\citet{zhang2025mirage} localizing the exact reasoning steps where hallucinations arise and follow-up diagnostic benchmarks~\citep{fan2026halluhard,liu2026agenthallu,zhan2026your} (summarized in Table~\ref{tab:taxonomy}).
Despite advances, existing efforts exhibit two fundamental shortcomings.
First, current methods rely on expert verification or coarse statistical metrics~\citep{manakul2023selfcheckgpt}, which can detect \emph{whether} a hallucination occurs but fail to explain \emph{why} it arises or trace it back to the specific reasoning step that produced it.
Second, existing open-world benchmarks transfer poorly to scientific domains, where densely interconnected dependencies allow a single erroneous claim to propagate through downstream reasoning and corrupt the entire output~\citep{li2023halueval}, a failure mode that flat, query-level evaluations cannot capture.
As our preliminary analysis (Section~\ref{find_and_dis}) reveals, scientific agent hallucinations diverge fundamentally from standard tasks: they are predominantly driven by a failure to master \emph{domain-specific priors} and their \emph{conceptual interconnectivity}.

To address these gaps, we introduce \textsc{Schema}, the first evidence-grounded framework that systematically traces and quantifies hallucinations in scientific agents.
\textsc{Schema} reframes evaluation from isolated query-level checks to topology-aware assessments grounded in structured domain knowledge, offering a domain-agnostic paradigm that generalizes across scientific disciplines.
To construct the benchmark, we develop an automated pipeline that integrates multi-source literature, LLMs, and external tools to build a highly reliable scientific concept graph.
Building on this graph, \textsc{Schema} generates a diverse suite of complex tasks ranging from multi-hop reasoning to programmatic experimental execution—going beyond prior datasets dominated by simplistic, single-hop queries.
This design explicitly probes both the breadth and depth of an agent's conceptual understanding, reflecting key stages of actual scientific discovery.
Our main contributions are summarized as follows:
\begin{itemize}
    \item \textbf{A Topology-Aware Evaluation Paradigm.} 
    We introduce \textsc{Schema}, the first evidence-grounded framework that reframes agent hallucination evaluation from isolated query-level checks to graph-grounded, multi-step trajectory assessment. 
    Built on a domain-agnostic pipeline, \textsc{Schema} moves beyond static benchmarking toward principled diagnosis.
    
    \item \textbf{Counterfactual Attribution for Causal Diagnosis.} 
    We develop a counterfactual reasoning method that automatically localizes hallucinations to the exact reasoning step where they originate and identifies their causal mechanism. 
    This shifts hallucination analysis from labor-intensive expert inspection to scalable, automated diagnosis.
    
    \item \textbf{A Domain-Specific Scientific Benchmark.} 
    We release a biomedical benchmark suite spanning two complementary subdomains of escalating complexity, supporting diverse task types from multi-hop reasoning to experimental execution, with concept importance grounded in graph topology.
    
    \item \textbf{Empirical Insights into Scientific Hallucinations.} 
    Through \textsc{Schema}, we reveal that scientific hallucinations stem primarily from flawed \emph{domain-specific priors} and weak \emph{conceptual interconnectivity}, with errors on topologically central concepts triggering cascading failures across reasoning chains. 
    These findings motivate a fundamental shift from surface-level accuracy metrics toward \emph{mechanism-level attribution}.
\end{itemize}

\section{Counterfactual Reasoning for Hallucination Detection and Attribution}
\label{sec:counterfactual}
Existing attribution methods rely heavily on costly and potentially subjective human evaluation~\citep{liu2026agenthallu, zhan2026your}.
To address this, we cast hallucination attribution in LLM agents as a \emph{counterfactual intervention} problem over ReAct-style~\citep{yao2022react} trajectories.
Our framework answers a causal question: \emph{Would the agent have produced the correct answer had the hallucinated content at step $t$ been eliminated?}

\begin{figure}[t]
    \centering
    \includegraphics[width=1\textwidth]{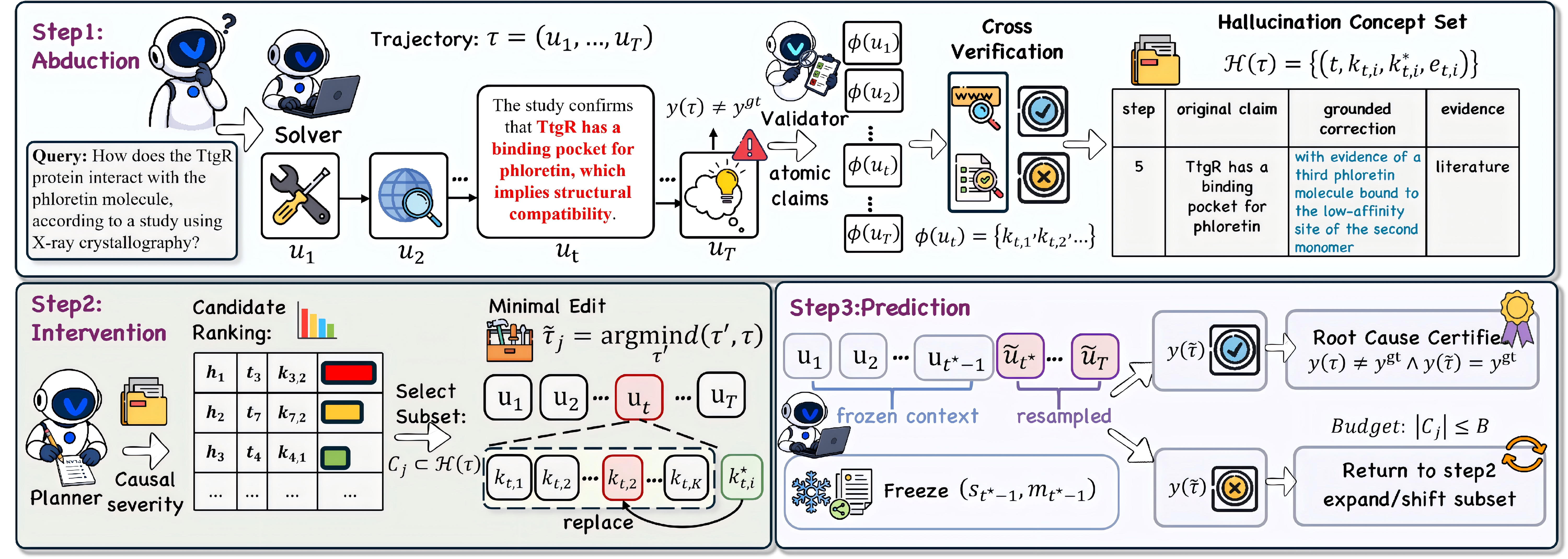}
    \caption{Our framework involves three stages: (i) \textbf{Abduction}: Trajectory generation and cross-verification to list hallucinations. (ii) \textbf{Intervention}: Iterative editing to replace false claims with grounded corrections. (iii) \textbf{Prediction}: Frozen-context resampling to certify the root cause.}
    \label{fig:tool_ovewview}
    \vspace{-10pt}
\end{figure}

\subsection{Counterfactual Attribution via Multi-Agent Collaboration}
To operationalize the above abstract concept within complex agentic workflows, we map the agent's sequential reasoning into a \emph{counterfactual attribution framework}, as shown in Figure~\ref{fig:tool_ovewview}.  
Our multi-agent system executes the three-step counterfactual evaluation procedure~\citep{pearl2009causality}—\textbf{Abduction}, \textbf{Intervention}, and \textbf{Prediction}—to systematically isolate the root causes of failure.

\noindent \textbf{Abduction}. A \emph{solver agent} $\pi_s$ generates a trajectory $\tau = (u_1, \ldots, u_T)$, where each step $u_t = (c_t, a_t, o_t)$ is tracked via AGDebugger~\citep{epperson2025interactive}.
If the final answer $y(\tau)$ diverges from the ground truth $y^{\mathrm{gt}}$, $\tau$ is flagged as a failure.
To verify hallucinations, a \emph{validator agent} $\pi_v$ extracts claims $\phi(u_t)$ from each step.
To prevent false positives from fabricated references, $\pi_v$ cross-verifies each claim against a source pool (literature and web).
Claims refuted by multiple independent sources form the \emph{hallucination concept set} $\mathcal{H}(\tau) = \{(t, k_{t, i}, k^{\star}_{t, i}, e_{t, i})\}$, recording the exact location, the original claim, its grounded correction, and the supporting evidence.

\noindent \textbf{Intervention}. To perform counterfactual probing, a \emph{planner agent} $\pi_p$ ranks candidates in $\mathcal{H}(\tau)$ by their potential structural impact and iteratively edits $\tau$.
At iteration $j$, for a selected subset $\mathcal{C}_j \subseteq \mathcal{H}(\tau)$, $\pi_p$ produces a minimally edited trajectory $\tilde\tau_j = \arg\min_{\tau'} d(\tau', \tau)$.
This edit ensures that hallucinated claims in $\mathcal{C}_j$ are replaced by their verified corrections while preserving all unselected content verbatim.
If the intervention fails to correct the outcome, the planner adaptively expands or shifts its candidate subset until a successful edit is found or the predefined attempt budget is exhausted.

\noindent \textbf{Prediction}. The solver $\pi_s$ resumes execution from the edited step $t^{\star}$. 
Crucially, leveraging the state-restoration capabilities of AGDebugger, we strictly freeze the prior context and memory state $(s_{t^{\star}-1}, m_{t^{\star}-1})$. 
Since the freezing is an architectural guarantee, it eliminates upstream sampling variance and strictly isolates the downstream impact of the intervention. 
Downstream units are then resampled as $\tilde u_{t} \sim \pi_s(\cdot \mid s_{t^{\star}-1}, m_{t^{\star}-1}, \tilde u_{t^{\star}:t-1})$.
A candidate subset $\mathcal{C}_j$ is certified as the root cause of the hallucination if and only if the rerun strictly flips the outcome ($y(\tau) \neq y^{\mathrm{gt}} \wedge y(\tilde\tau) = y^{\mathrm{gt}}$).

\subsection{Experimental Setups}
We implement our framework using Intern-S1-Pro~\citep{zou2026intern} as the agent model, with AutoGen~\citep{wu2024autogen} and AGDebugger~\citep{epperson2025interactive} serving as the orchestration framework.
For domain-specific evaluation, we randomly sample $346$ questions from ProteinLMBench~\citep{shen2024fine}, isolating the $141$ failure trajectories for counterfactual analysis.

\subsection{Findings and Discussions}
\label{find_and_dis}

\paragraph{Finding 1: Hallucinations Exhibit Distinct Modes with Divergent Repairability.}
Counterfactual analysis decomposes failures into three structural categories: \texttt{wrong\_answer}, \texttt{tool\_loop}, and \texttt{no\_answer}.
The dominant \texttt{wrong\_answer} mode is driven primarily by \emph{mapping} and \emph{fact} errors.
\emph{Mapping} errors—where agents retrieve the correct evidence but fail to logically connect it to the final answer—account for the majority of hallucinations yet remain comparatively recoverable.
In contrast, \emph{fact} errors exhibit the lowest repairability: once an agent commits to a false premise, subsequent steps inherit and amplify the misconception.
This asymmetry indicates that hallucination severity is governed not by surface frequency but by the position and type of error within the reasoning chain.



\begin{wrapfigure}{r}{0.5\textwidth}
  \centering
  \fontsize{9pt}{9pt}\selectfont

  \captionof{table}{Hallucination taxonomy and repair outcomes on the Protein-QA subset.}\label{tab:taxonomy}\vspace{-5pt}
  \setlength{\tabcolsep}{3pt}
  \begin{tabular}{lrrr}
    \toprule
    \textbf{Category} & \textbf{\#Total} & \textbf{\#Repaired} & \textbf{Fix rate} \\
    \midrule
    \texttt{wrong\_answer} & 110 & 52 & 47.3\,\% \\
    \quad $\hookrightarrow$ \emph{mapping\_error}    & 68 & 33 & 48.5\,\% \\
    \quad $\hookrightarrow$ \emph{fact\_error}       & 36 & 13 & 36.1\,\% \\
    \quad $\hookrightarrow$ \emph{alignment\_error}  & 4  & 4  & 100\,\% \\
    \quad $\hookrightarrow$ \emph{constraint\_error} & 2  & 2  & 100\,\% \\
    \texttt{tool\_loop} & 29 & 12 & 41.4\,\% \\
    \texttt{no\_answer} & 2  & 1  & 50.0\,\% \\
    \midrule
    \textbf{Total} & \textbf{141} & \textbf{65} & \textbf{46.1\,\%} \\
    \bottomrule
  \end{tabular}

  \vspace{2pt}
    
  \includegraphics[width=\linewidth]{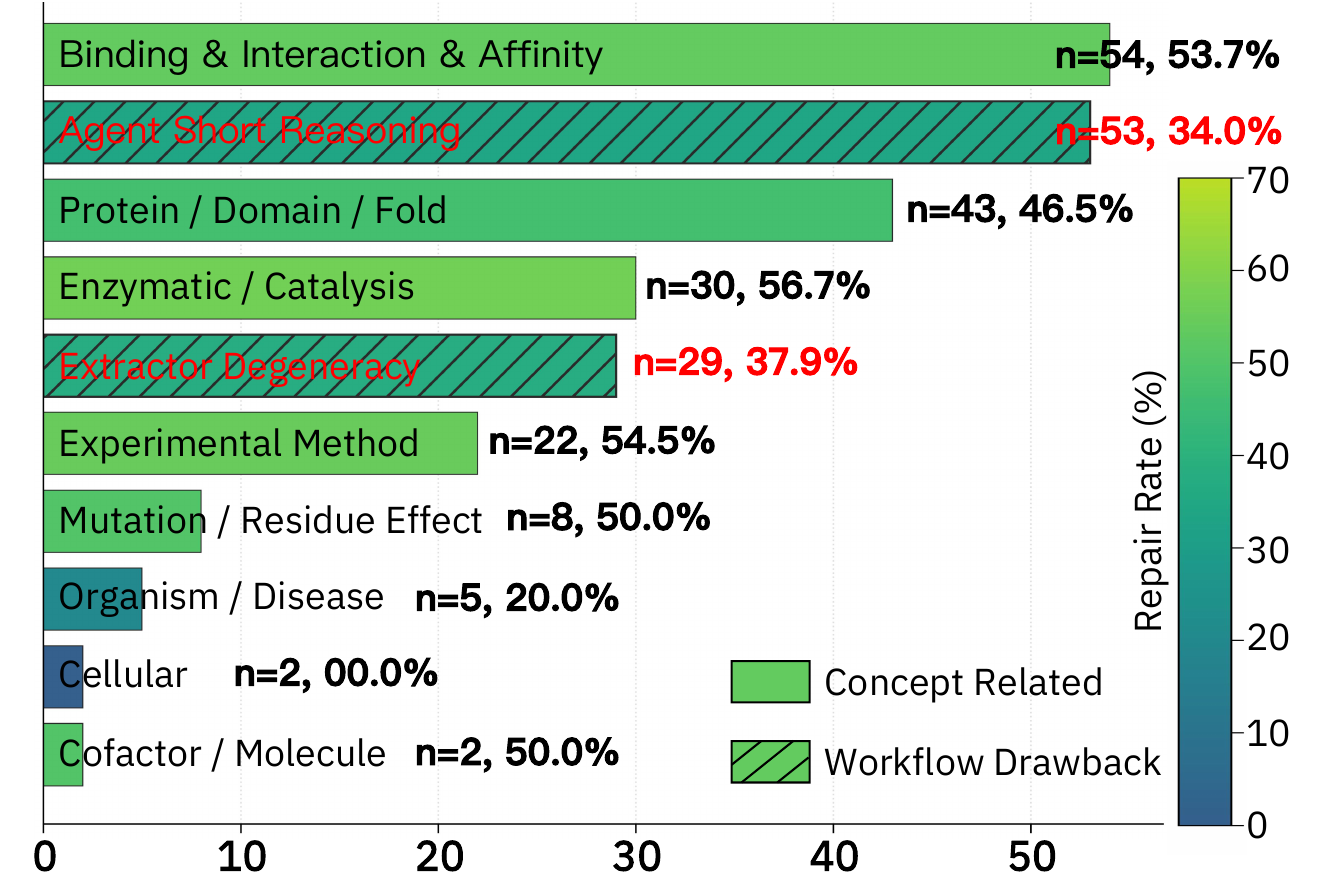}
  \captionof{figure}{Distribution of the 248 hallucinated concepts across semantic themes on the Protein-QA subset.
  End-of-bar annotations report the count $n$ and the corresponding repair rate (\%).}\label{fig:theme_dist}
  \vspace{-15pt}
\end{wrapfigure}

\paragraph{Finding 2: Hallucinated Claims Cluster around a Few Knowledge Hubs.}
Figure~\ref{fig:theme_dist} reveals the \emph{conceptual} origins of failures.
While roughly one-third of failures stem from workflow-level issues such as truncated reasoning and extractor degeneracy, the majority (67\%) trace back to \emph{flawed domain knowledge}.
Critically, these knowledge errors are far from uniformly distributed: they concentrate around a small set of recurring concept hubs, and \textsc{Enzymatic/Catalysis} alone accounts for nearly half of all hallucinations—accompanied by a long tail of sporadic concepts each contributing fewer than five cases.
This hub-centric distribution suggests that hallucinations in scientific agents are not random lapses but systematic blind spots tied to specific high-traffic concepts.
Notably, repair rates further reinforce this structural view: hub concepts remain comparatively recoverable, whereas long-tail rare concepts collapse to near-zero repairability, indicating that error severity is governed less by frequency than by where a concept sits in the knowledge topology.

\paragraph{Discussion: Bridging Failure Modes to Evaluation Design.}
Our findings motivate two design principles for scientific-agent benchmarks~\citep{liu2026agenthallu, zhan2026your}.
First, the existence of distinct error modes with divergent repairability implies that a single accuracy score cannot diagnose where an agent actually breaks down.
\emph{Mapping}, \emph{fact}, and workflow-level failures arise from fundamentally different reasoning bottlenecks, and each must be elicited by a corresponding task design.
This calls for \textbf{stratified, multi-format evaluation} that pairs distinct question types with the reasoning capability they target, rather than collapsing heterogeneous failures into one number.
Second, the hub-centric distribution of hallucinations reveals that not all factual errors carry equal weight: a small set of high-traffic concepts dominates failure frequency, while sparse peripheral concepts dominate failure severity.
A faithful evaluation must therefore be \textbf{topology-aware}, weighting concepts by their structural role rather than treating all facts as interchangeable.
Taken together, these principles point toward a shift in agent evaluation—from aggregate accuracy scores toward a finer-grained examination of where errors originate, how they propagate, and which concepts they cluster around.

\section{\textsc{Schema}: Diagnosing Scientific Hallucinations via Domain Ontologies}

To address the aforementioned challenges, we introduce \textsc{Schema} (\textbf{S}cientific \textbf{C}oncept \textbf{H}ierarchies and \textbf{E}vidence-grounded \textbf{M}etrics for \textbf{A}gents).
Moving beyond surface-level accuracy, \textsc{Schema} is architected to reflect the structural realities of scientific reasoning.


\subsection{Overview of \textsc{Schema}}
\begin{figure}[h]
    \centering
    \includegraphics[width=1\textwidth]{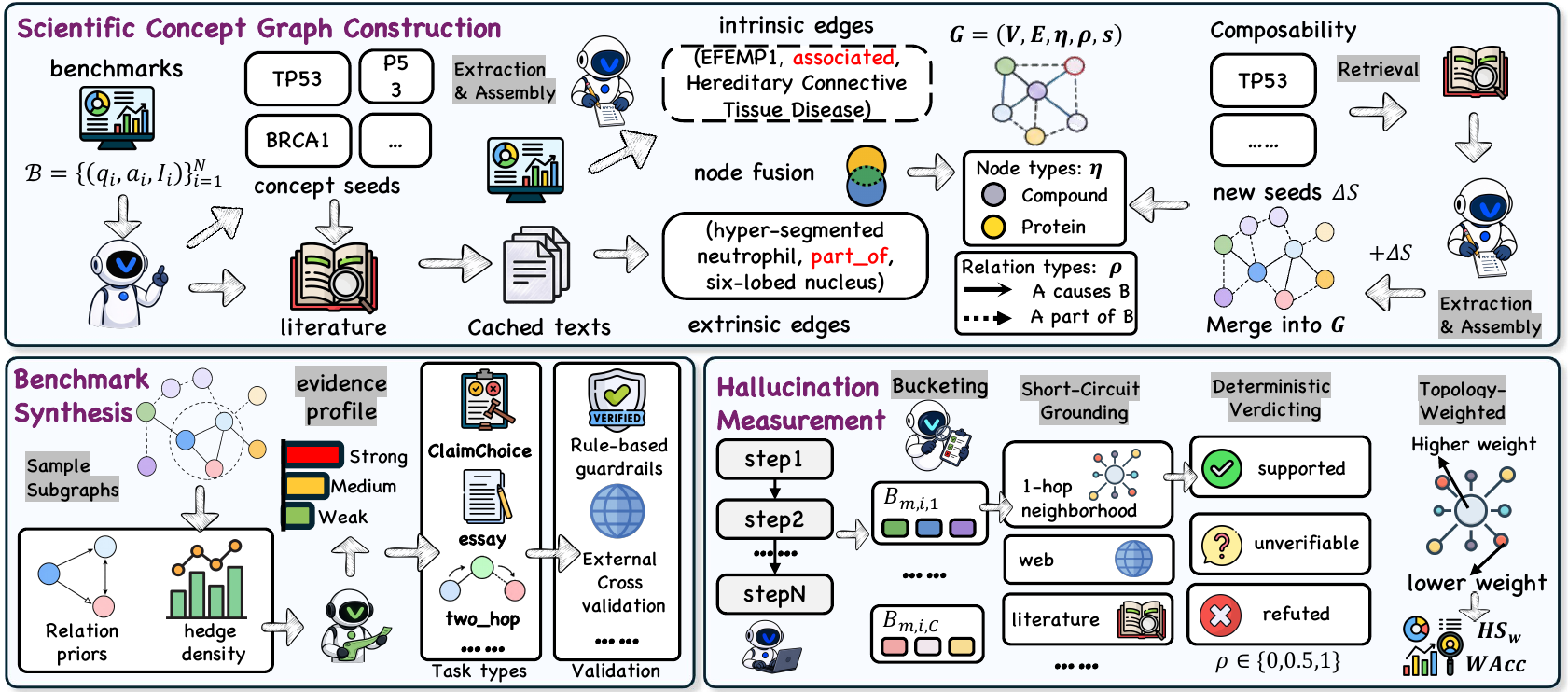}
    \caption{Overview of the \textsc{Schema} pipeline. 
    The framework comprises three stages: concept-graph construction, evidence-grounded benchmark synthesis, and topology-aware hallucination evaluation.}
    \label{fig:evaluation_ovewview}
    \vspace{-10pt}
\end{figure}

\begin{figure*}[t]
\centering
\begin{minipage}[t]{0.54\textwidth}
    \centering
    \vspace{0pt}  
    \includegraphics[width=\linewidth]{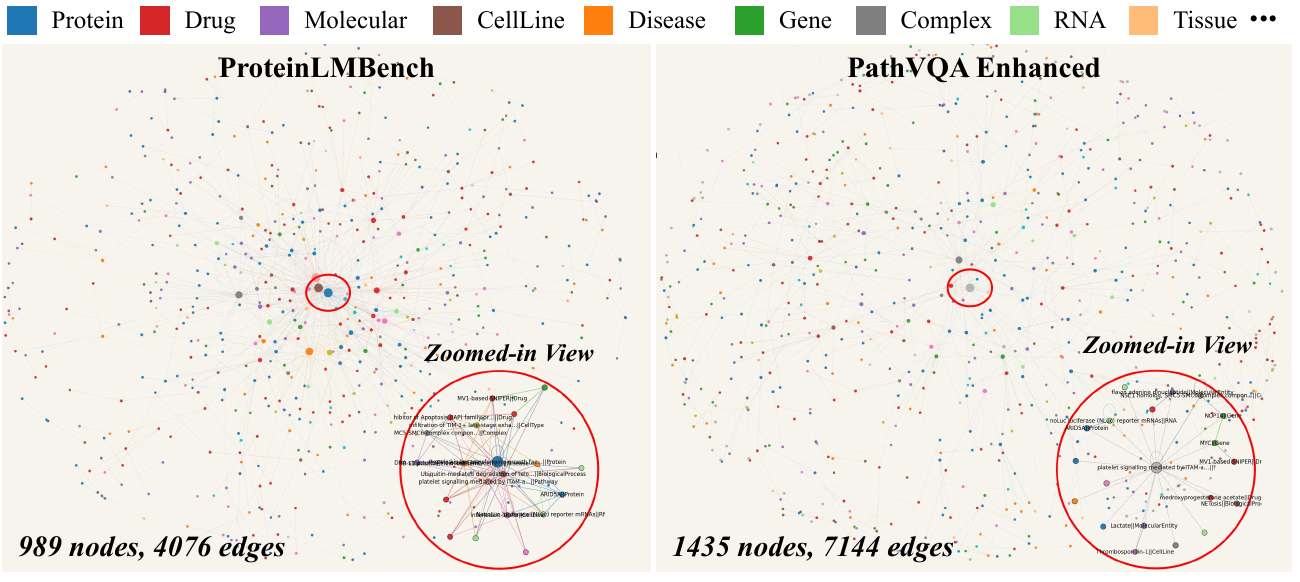}
    \captionof{figure}{Visualization of Protein Graph (left) and its evolved version PathVQA-Enhanced (right). 
    Zoomed-in views highlight local connectivity around hub nodes.
    }
    \label{fig:graph-vis}
\end{minipage}%
\hfill
\begin{minipage}[t]{0.44\textwidth}
    \centering
    \vspace{0pt}  
    \captionof{table}{Question-type $\times$ importance-tier cross-tabulation. \textbf{Core}/\textbf{Peri.}/\textbf{Off} denote node-importance tiers from graph centrality.}
    \label{tab:qtype-tier}
    \scriptsize
    \setlength{\tabcolsep}{2.5pt}
    \renewcommand{\arraystretch}{1.1}
    \definecolor{tierCore}{HTML}{2C3E50}
    \definecolor{tierPeri}{HTML}{5D6D7E}
    \definecolor{tierOff}{HTML}{95A5A6}
    \definecolor{rowShade}{HTML}{F4F6F7}
    \begin{tabular}{l ccc c ccc}
    \toprule
    \multirow{2}{*}{\textbf{Q-Type}}
    & \multicolumn{3}{c}{\textbf{Protein (800)}} & & \multicolumn{3}{c}{\textbf{PathVQA (+495)}} \\
    \cmidrule(lr){2-4} \cmidrule(lr){6-8}
    & {\color{tierCore}\textbf{Core}} 
    & {\color{tierPeri}\textbf{Peri.}} 
    & {\color{tierOff}\textbf{Off}} 
    & 
    & {\color{tierCore}\textbf{Core}} 
    & {\color{tierPeri}\textbf{Peri.}} 
    & {\color{tierOff}\textbf{Off}} \\
    \midrule
    \texttt{Claim}    & 128 & 151 & 103 & & 62 & 62 & 74 \\
    \rowcolor{rowShade}
    \texttt{Essay}    & 57  & 117 & 65  & & 51 & 64 & 54 \\
    \texttt{Bool.}    & 13  & 51  & 44  & & 12 & 37 & 38 \\
    \rowcolor{rowShade}
    \texttt{2-Hop}    & 20  & 2   & 25  & & 4  & 1  & 15 \\
    \texttt{Exp.}     & 18  & 1   & 5   & & 15 & 2  & 4  \\
    \midrule
    \textbf{Total}    & \textbf{236} & \textbf{322} & \textbf{242} & & \textbf{144} & \textbf{166} & \textbf{185} \\
    \bottomrule
    \end{tabular}
\end{minipage}
\vspace{-10pt}
\end{figure*}

\paragraph{Description.}
To demonstrate the scalability of our evaluation framework, we instantiate \textsc{Schema} in the biomedical domain, where the dense interconnections among proteins, pathways, and diseases provide a natural testbed for graph-based analysis (Figure~\ref{fig:evaluation_ovewview}).
Within this domain, we apply our automated graph-construction pipeline to two representative subfields of increasing structural complexity, allowing us to assess the framework along two complementary axes: depth of foundational knowledge and breadth of interactive reasoning.
We first instantiate \textsc{Schema} on the \textsc{Protein Domain} to probe molecular-level priors, and then extend it to \textsc{PathVQA-Enhanced}~\citep{he2020pathvqa} to capture broader disease-level interactions.
This progressive design shows that our topology-aware pipeline transfers naturally across subfields of differing granularity, and yields a benchmark in which concept importance is explicitly quantified by topology rather than assumed uniform (Figure~\ref{fig:graph-vis}, Table~\ref{tab:qtype-tier}).

\subsection{Scientific Concept Graph Construction}\label{sec:method-graph-construction}
\noindent \textbf{Problem Setup.} 
To bypass costly manual annotation, we derive concept seeds directly from human-verified benchmarks.
Formally, given a benchmark $\mathcal{B} = \{(q_i, a_i, I_i)\}_{i=1}^N$, we construct a graph
\begin{equation}
\mathcal{G} = (\mathcal{V}, \mathcal{E}, \eta, \rho, s), \quad 
\eta: \mathcal{V} \to \mathcal{T}, \quad 
\rho: \mathcal{E} \to \mathcal{R}, \quad 
s: \mathcal{E} \to \{\textit{intrinsic}, \textit{extrinsic}\},
\label{eq:graph-def}
\end{equation}
where $\mathcal{T}$ and $\mathcal{R}$ denote closed sets of biomedical entity types and canonical relations, and the provenance function $s$ records the origin of each edge.
\textit{Intrinsic} edges are extracted from the internal signal of $\mathcal{B}$ itself—answer text, accompanying annotations, and question structure—capturing the human-curated expertise.
\textit{Extrinsic} edges are generated by expanding outward from the seeds: we retrieve from peer-reviewed literature databases (\textit{e.g.}, PubMed, bioRxiv) and web-based scientific dialogue, extracting additional relations that connect concepts beyond any single benchmark.

\noindent \textbf{Construction Pipeline.} 
The graph is assembled through a three-stage pipeline; full implementation details are provided in Appendix~\ref{app:concept_graph}.
\textbf{(1) Seed Extraction \& Retrieval.} 
Candidate biomedical terms are obtained via a cascaded rule-based and LLM-based extractor, then used to query multi-source literature databases (PubMed, bioRxiv, etc.).
Retrieved papers are ranked by a combination of semantic similarity and venue impact.
\textbf{(2) Triple Extraction \& Node Fusion.} 
An LLM extracts knowledge triples from cached texts.
To consolidate fragmentary nodes, we fuse pairs that exceed an embedding-similarity threshold \emph{only when they share the same type under $\eta$}, preventing semantically close but ontologically distinct entities (\textit{e.g.}, a protein and its encoding gene) from being collapsed.
The same fusion step also aligns \textit{intrinsic} nodes from $\mathcal{B}$ with their \textit{extrinsic} counterparts in the literature graph.
\textbf{(3) Incremental Growth ($\oplus$).} 
The pipeline is designed for incremental growth rather than rebuilding from scratch.
When a new benchmark $\mathcal{B}'$ is introduced, retrieval and extraction are triggered only on the seed delta $\Delta\mathcal{S}$:
\begin{equation}
\mathcal{G} \oplus \Delta\mathcal{S} \,\to\, \mathcal{G}',
\label{eq:graph-growth}
\end{equation}
where the existing graph $\mathcal{G}$ is extended with new edges and nodes and merged via the same fusion procedure.
This growth operation scales linearly with $|\Delta\mathcal{S}|$ in retrieval and extraction, making \textsc{Schema} suitable for the open-ended expansion characteristic of scientific domains.

\subsection{Evidence-Grounded Benchmark Synthesis}\label{sec:question-gen}

Building upon $\mathcal{G}$, we synthesize an \emph{evidence-grounded} benchmark whose questions are anchored to verifiable literature evidence rather than bare relation triples.

\noindent \textbf{Topology-Aware Formulation.} 
For each sampled subgraph, we compute a discrete \emph{evidence profile} $s \in \{\textsc{weak}, \textsc{medium}, \textsc{strong}\}$ that aggregates three signals: a per-relation strength prior, the density of hedging language in supporting text, and the number of independent supporting sources per hop (full decision rule in Appendix~\ref{app:question_generation}).
This profile serves a dual role: it \emph{gates} which question types a subgraph is eligible for, and it \emph{modulates} claim phrasing so that wording remains calibrated to the underlying evidence.
The resulting suite spans five complementary task types covering distinct cognitive operations—\textsc{ClaimChoice}, \textsc{BooleanSupport}, \textsc{TwoHopTail}, \textsc{Essay}, and \textsc{ExperimentCode}—ranging from discriminative claim selection and binary support judgment to multi-hop reasoning, essay, and executable experimental code (detailed in Appendix~\ref{app:question_generation}).

\noindent \textbf{Validation Stack.} 
Generated candidates pass through a three-tier validation pipeline before entering the benchmark.
\textit{(i) Rule-based guardrails} enforce structural constraints across all types, including evidence support (the cited claim must overlap with its source chunk), multi-source confirmation (each claim must appear in at least two distinct documents), and answer uniqueness for selection-style questions.
\textit{(ii) Sandbox execution} validates \textsc{ExperimentCode} programmatically: the reference solution must pass all unit tests while the masked variant must fail at least one, ensuring that the blanked content is load-bearing rather than cosmetic.
\textit{(iii) LLM-as-Judge cross-validation} re-examines surviving text candidates by checking whether the claim is consistently supported across an evidence bundle assembled from multiple independent passages and live literature retrieval (top-$k$); a candidate is accepted only if the judge verdict is \textsc{supported}, with \textsc{contradicted} or \textsc{insufficient} verdicts triggering rejection (judge configuration and reproducibility settings in Appendix~\ref{app:question_generation}).

\subsection{Topology-Aware Metrics for Hallucination Measurement}\label{sec:metrics}

To rigorously evaluate scientific agents, we decouple surface-level success from reasoning reliability. 
Beyond standard terminal accuracy, our \emph{trajectory hallucination pipeline} systematically audits an agent's intermediate reasoning steps $\tau_{m,i}$, producing a fine-grained severity metric that supports detection at scale.
This pipeline is distinct from the multi-agent counterfactual attribution module (Section~\ref{sec:counterfactual}), which performs deeper but more costly causal localization on a smaller subset of trajectories.
The pipeline measures hallucinations along two complementary dimensions: \textbf{claim-level coverage}, which ensures that every intermediate assertion is verified, and \textbf{topology-aware severity}, which weights each verified claim by its structural role in the knowledge graph.

\noindent\textbf{Claim-Level Coverage.} 
\textit{(1) Claim Extraction.} An extractor LLM canonicalizes and de-hedges factual claims across the entire multi-step trajectory, grouping them into per-concept buckets $\mathcal{B}_{m,i,c}$.
\textit{(2) Hierarchical Grounding.} Each claim is verified against an evidence package $\mathcal{E}_{m,i,c}$ assembled by querying four sources in cascade—gold chunks, the 1-hop graph neighborhood, web search, and live literature retrieval—and terminating at the first non-empty layer.
This short-circuit ordering is a deliberate efficiency choice that allows the pipeline to scale across long trajectories without exhausting retrieval budgets.
\textit{(3) Discrete Verdicting.} A \emph{separate} judge LLM assigns each claim a categorical verdict and maps it to a severity penalty $\rho \in \{0,\, 0.5,\, 1\}$ corresponding to \textsc{supported}, \textsc{unverifiable}, and \textsc{refuted}.
The discrete mapping is a design choice: it sidesteps the spurious precision of LLM-emitted continuous scores while preserving an ordinal severity signal.
Full prompts and tie-breaking rules are deferred to Appendix~\ref{app:hallu_pipeline}.

\noindent\textbf{Topology-Aware Severity.} 
Standard hallucination metrics treat all factual mistakes as equivalent. 
This is misleading in scientific reasoning: an error at a highly connected concept propagates through every downstream chain that passes through it, whereas an error at a peripheral concept remains localized.
We capture this asymmetry by assigning each claim a graph-based weight:
\begin{equation}
w_g(c) \;=\; 1 + \mathbb{1}\!\left[\eta(c) \neq \bot\right] \cdot \log\!\bigl(1 + \deg_{\mathcal{K}}(\eta(c))\bigr),
\label{eq:wg}
\end{equation}
where $\eta(c)$ maps the claim to its corresponding node in the global knowledge graph $\mathcal{K}$ via lexical matching with an embedding-based fallback, and $\deg_{\mathcal{K}}(\cdot)$ denotes the undirected node degree—directionality is treated as a property of each claim's verdict $\rho$, rather than of the underlying concept's centrality.
The weight is additive: off-graph claims fall back to the baseline $w_g = 1$, and in-graph claims receive a log-scaled bonus that grows with their structural connectivity but never lets a single hub dominate the metric.
We then aggregate per-claim severities under $w_g$ to obtain the \emph{Topology-Weighted Hallucination Severity} ($\mathrm{HS}^w$), and report it alongside the unweighted $\mathrm{HS}$ so that any shift between the two reflects the structural distribution of errors rather than their raw frequency; aggregation formulas and the rationale for the off-graph baseline are detailed in Appendix~\ref{app:metrics}.
Thus, $w_g$ measures the \emph{leverage} of an error—how far its consequences propagate through the knowledge graph—an aspect orthogonal to whether the error itself is easy or hard to repair.

\section{Experiments}
\subsection{Experimental Setup} 

\noindent \textbf{Baselines and Agent Workflow.}
We evaluate a diverse cohort of 10 contemporary instruction-tuned LLMs.
To isolate the models' intrinsic reasoning and tool-use capabilities, we deliberately exclude static retrieval-augmented generation baselines. 
Instead, all models operate within identical autogen-based workflows (a reasoning agent and a passive tool executor) equipped with literature search, web search, and other tools, detailed in Appendix~\ref{app:experimental_setup}.
Each sample is executed as an independent multi-turn dialogue with strict budget limits, and the full trajectory is persisted for hallucination analysis.

\noindent \textbf{Evaluation Metrics.}
We report two complementary families of metrics~\ref{app:metrics}. First, \emph{Terminal Accuracy} scores the final answer against a gold reference using a pairwise LLM judge (\textsc{GPT-4o}). 
Second, our contribution, the \emph{Hallucination Metric}, evaluates the intermediate reasoning steps.

\begin{table*}[h]
    \centering
    \caption{Main results on \textsc{Schema}. We report accuracy (\textbf{Acc}, 
    \textbf{WAcc}; higher is better) and three hallucination metrics 
    (\textbf{HR}, \textbf{HS}, \textbf{HS}$^{w}$; lower is better) across 10 LLMs on the Protein Domain and 
    PathVQA-Enhanced subsets. The \textbf{best} per column is 
    bolded and the \underline{second-best} is underlined.}
    \label{tab:main-results}
    \vspace{-5pt}
    \begin{center}
        \fontsize{8.5pt}{10pt}\selectfont
        \setlength{\tabcolsep}{4pt}
        \renewcommand{\arraystretch}{1.15}
        \begin{tabular}{l @{\hskip 4mm} ccccc @{\hskip 6mm} ccccc}
        \toprule
         \multirow{2}{*}{\textbf{Models}} 
         & \multicolumn{5}{c}{\textbf{Protein Domain} ($n{=}800$)} 
         & \multicolumn{5}{c}{\textbf{PathVQA-Enhanced} ($n{=}495$)} \\
         \cmidrule(lr){2-6} \cmidrule(lr){7-11}
         & \textbf{Acc}$\uparrow$ & \textbf{WAcc}$\uparrow$ 
         & \textbf{HR}$\downarrow$ & \textbf{HS}$\downarrow$ 
         & \textbf{HS}$^{w}\downarrow$
         & \textbf{Acc}$\uparrow$ & \textbf{WAcc}$\uparrow$ 
         & \textbf{HR}$\downarrow$ & \textbf{HS}$\downarrow$ 
         & \textbf{HS}$^{w}\downarrow$ \\
         \midrule
         \multicolumn{11}{l}{\textit{Open-source LLMs}} \\
         \midrule
         Kimi-K2.5              & 0.299 & 0.371 & \textbf{0.050} & 0.275 & 0.309 & 0.595 & 0.586 & 0.096 & 0.360 & 0.375 \\
         Llama-4-Scout          & 0.600 & 0.602 & 0.085 & 0.273 & 0.296 & 0.588 & 0.577 & 0.059 & 0.280 & \underline{0.291} \\
         Doubao-Seed-2.0-Pro    & \underline{0.739} & \underline{0.740} & 0.083 & 0.283 & 0.303 & \textbf{0.677} & \textbf{0.662} & \underline{0.017} & 0.247 & 0.407 \\
         Intern-S1-Pro          & 0.428 & 0.441 & 0.061 & \underline{0.206} & \textbf{0.262} & 0.489 & 0.475 & 0.125 & 0.322 & 0.343 \\
         DeepSeek-V4-Flash      & 0.591 & 0.584 & 0.068 & 0.353 & 0.362 & \underline{0.675} & \underline{0.662} & 0.080 & 0.394 & 0.408 \\
         \midrule
         \multicolumn{11}{l}{\textit{Closed-source API LLMs}} \\
         \midrule
          Qwen3.6-Plus           & 0.546 & 0.586 & \underline{0.056} & 0.242 & 0.277 & 0.555 & 0.568 & \textbf{0.009} & 0.256 & 0.432 \\
         GPT-4o                 & 0.482 & 0.487 & 0.132 & 0.270 & 0.290 & 0.418 & 0.411 & 0.019 & 0.171 & 0.323 \\
         GPT-5.4-mini           & 0.592 & 0.605 & 0.086 & \textbf{0.179} & \underline{0.268} & 0.609 & 0.594 & 0.024 & \textbf{0.122} & \textbf{0.287} \\
         Gemini-3-Flash & \textbf{0.747} & \textbf{0.748} & 0.068 & 0.297 & 0.326 & 0.529 & 0.530 & 0.048 & 0.274 & 0.326 \\
         Grok-4.1-Fast          & 0.554 & 0.544 & 0.078 & 0.187 & 0.352 & 0.658 & 0.635 & 0.021 & \underline{0.163} & 0.434 \\
         \bottomrule
        \end{tabular}
    \end{center}
    \vspace{-15pt}
\end{table*}

\subsection{Main Results}
\begin{figure}[t]
    \centering
    \includegraphics[width=1.0\linewidth]{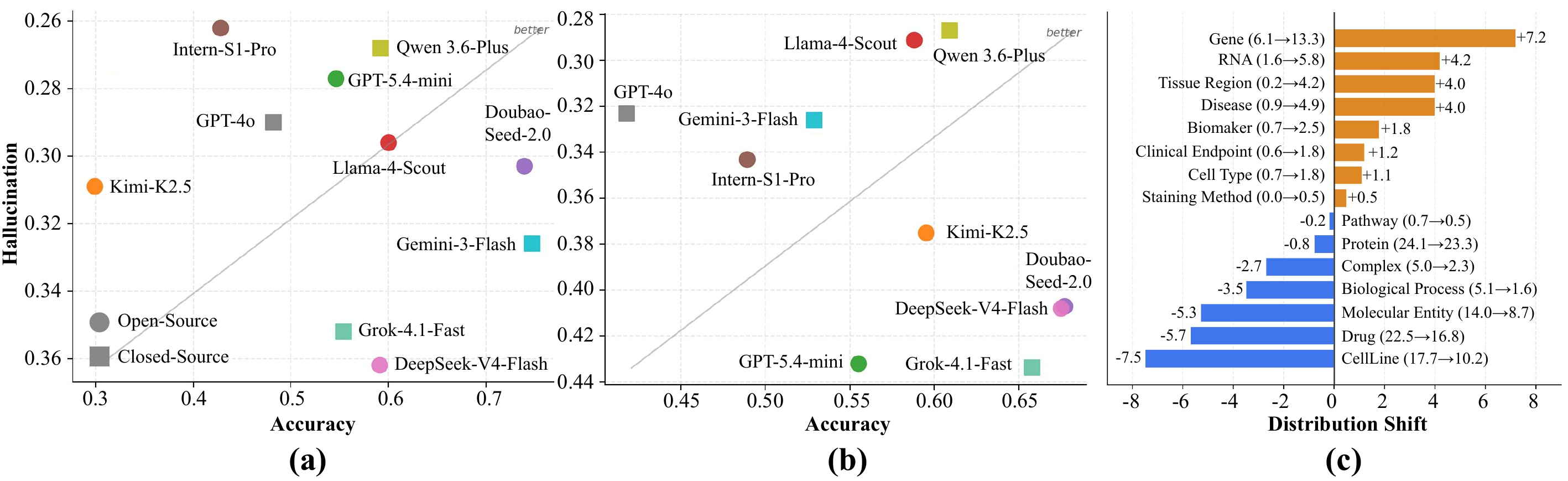}
     \caption{(a, b) Final-answer accuracy versus weighted hallucination severity HS$^{w}$ on the Protein Domain (left) and PathVQA-Enhanced (right).
     Better models sit toward the upper-right; circles denote open-source models, squares denote closed-source APIs. 
     (c) The two benchmarks operate on near-disjoint concept spaces: the diverging bar shows the $\Delta$ shift per concept type.}
    \label{fig:acc_hallu}
    \vspace{-15pt}
\end{figure}

\paragraph{Capability Bottlenecks under Concept Shift.}
Figure~\ref{fig:acc_hallu}(c) confirms that our two subsets operate on near-disjoint concept spaces, shifting from foundational molecular entities in the Protein Domain to highly interconnected clinical pathways in PathVQA-Enhanced.
Table~\ref{tab:main-results} reveals how this topological shift exposes distinct capability bottlenecks.
On the Protein Domain, proprietary models such as Gemini-3-Flash dominate, with accuracy spanning a wide 0.30–0.75 range that suggests heavy reliance on memorizing static domain priors.
When the distribution shifts to complex pathway interactions, however, proprietary dominance fades and open-source models (Doubao-Seed-2.0-Pro, DeepSeek-V4-Flash) take the lead, nearly halving the overall performance gap.
This indicates that retrieving isolated foundational facts and reasoning over complex concept interactions demand fundamentally different model capabilities.

\paragraph{Topological Robustness and Error Isolation.}
Switching from accuracy to hallucination evaluation reveals a clear divide in how models manage reasoning failures.
Several open-source models achieve impressively low raw hallucination rates (HR)—Qwen3.6-Plus and Doubao-Seed-2.0-Pro lead on PathVQA-Enhanced—yet struggle to contain the \emph{topological severity} of these errors, registering some of the highest HS$^{w}$ values in the benchmark (0.432 and 0.407, respectively).
Closed-source models, in contrast, occupy the Pareto frontier of structural reliability: GPT-5.4-mini achieves the lowest base severity (HS) on both subsets and the lowest HS$^{w}$ on PathVQA-Enhanced (0.287), confining its errors to peripheral concepts rather than central scientific hubs.
This validates our topology-weighted metric: counting hallucinations alone is insufficient, as the structural location of an error determines how far its consequences propagate.

\paragraph{Decoupling of Accuracy and Trajectory Honesty.}
Figure~\ref{fig:acc_hallu}~(a, b) reveals that final accuracy and topology-weighted hallucination severity ($\mathrm{HS}^w$) are only weakly correlated, exposing a striking decoupling: models frequently reach correct terminal answers via flawed intermediate logic—a form of \emph{shortcut reasoning}.
For instance, DeepSeek-V4-Flash and GPT-5.4-mini tie on Protein Domain accuracy (0.59), yet differ sharply in trajectory quality ($\mathrm{HS}^w$ of 0.362 vs. 0.268).
Similarly, Gemini-3-Flash leads on accuracy but shows middling trajectory honesty, whereas Intern-S1-Pro maintains a low $\mathrm{HS}^w$ despite weaker accuracy.
Because this disconnect persists across two near-disjoint concept spaces, it is a fundamental characteristic of LLM agents rather than a corpus-specific artifact.
This finding underscores the necessity of our trajectory-level evaluation: judging high-stakes scientific agents on terminal answers alone masks silent reasoning failures upstream.

\subsection{More Discussion about \textsc{Schema}}

\paragraph{Refutation Complexity Scales Topological Severity.}
To move beyond the colloquial intuition that models ``remember names but fabricate relations,'' we organize concepts along a \emph{verifiability axis} that orders them by the cognitive cost of refutation: from single-step database lookups (Identifier-grounded) and multi-component matching (Compositional) to multi-step causal tracing (Processual) and original methodological verification (Locative).

\begingroup
\setlength{\columnsep}{12pt}
\setlength{\intextsep}{6pt}
\begin{lrbox}{\schemaconceptfigurebox}
    \begin{minipage}{0.47\textwidth}
    \centering
    \vspace{-8pt}
    \includegraphics[width=0.96\linewidth]{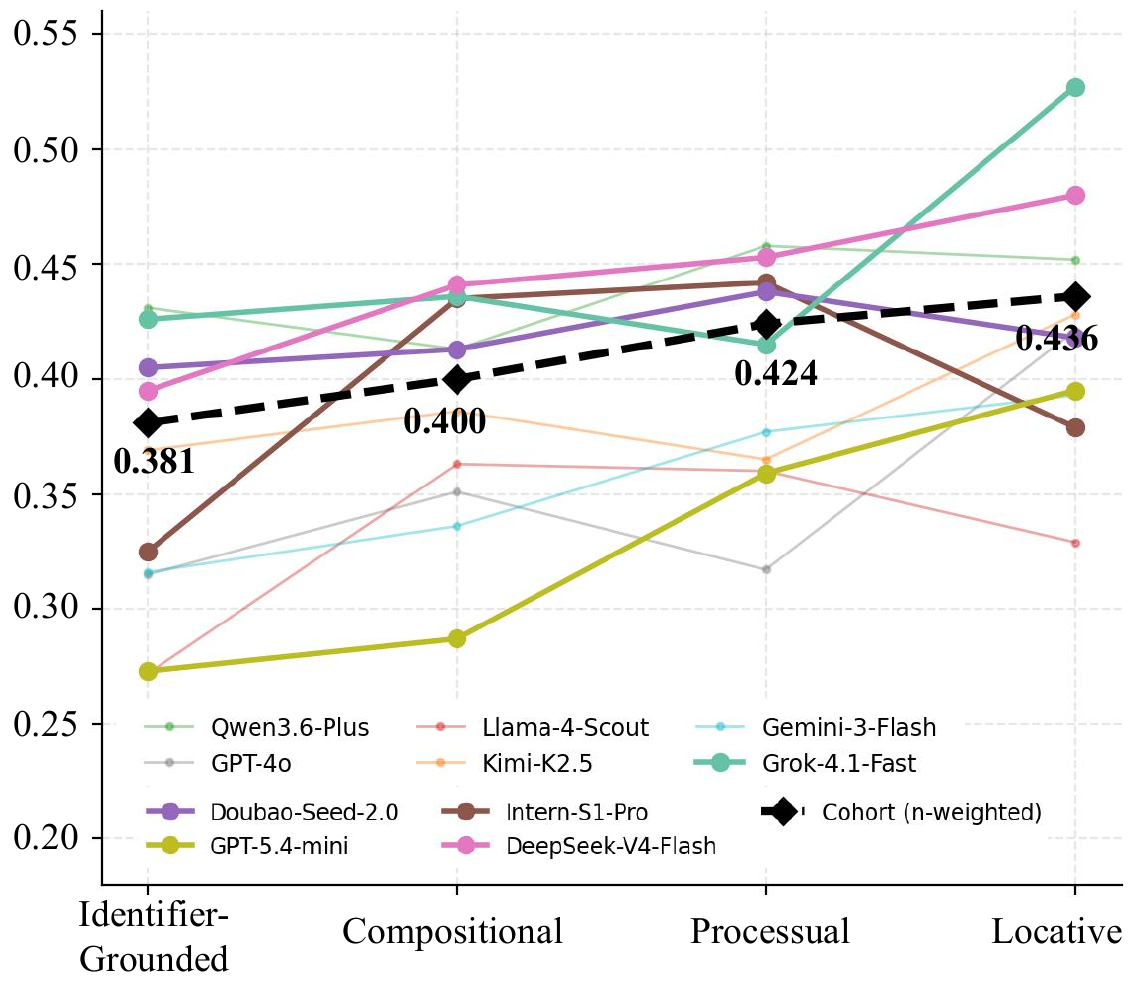}
    \vspace{-3pt}
    \captionof{figure}{Topology-weighted hallucination severity ($\mathrm{HS}^w$) along the verifiability axis on the Protein Domain.
    We aggregate 15 fine-grained concept types into four functional categories ordered by verification complexity.
    The dashed black line indicates the sample-weighted cohort mean.}
    \label{fig:concept_detail}
    \vspace{-4pt}
    \end{minipage}
\end{lrbox}
\Needspace{\dimexpr\ht\schemaconceptfigurebox+\dp\schemaconceptfigurebox
    +2\intextsep+2\baselineskip\relax}
\begin{wrapfigure}{r}[0pt]{0.47\textwidth}
    \usebox{\schemaconceptfigurebox}
\end{wrapfigure}
\noindent
As Figure~\ref{fig:concept_detail} shows, the cohort-averaged topology-weighted severity ($\mathrm{HS}^w$) rises monotonically along this axis—from $0.381$ at Identifier-grounded to $0.436$ at Locative—indicating that hallucination severity scales with how hard a claim is to refute.
The axis also exposes a clear decoupling between error frequency and structural impact: models often fabricate identifier-grounded entities (highest HR, lowest $\mathrm{HS}^w$) but commit far less frequently to processual mechanisms, where each fabrication triggers more severe causal failures.
Per-model patterns reinforce this view, with the largest divergences appearing at the high-complexity end of the axis: \textsc{Grok-4.1-Fast} concentrates its failures on locative concepts, where its $\mathrm{HS}^w$ spikes to roughly $0.53$ (the highest in the cohort), while \textsc{Qwen3.6-Plus} stays below the cohort average across the first three categories but jumps to $\sim\!0.48$ at Locative.
Together, these patterns suggest that as concepts move toward higher refutation complexity, model differences amplify and structurally critical hallucinations become harder to contain.
\par
\WFclear
\endgroup

\begin{figure*}[t]
    \centering
    \includegraphics[width=0.9\linewidth]{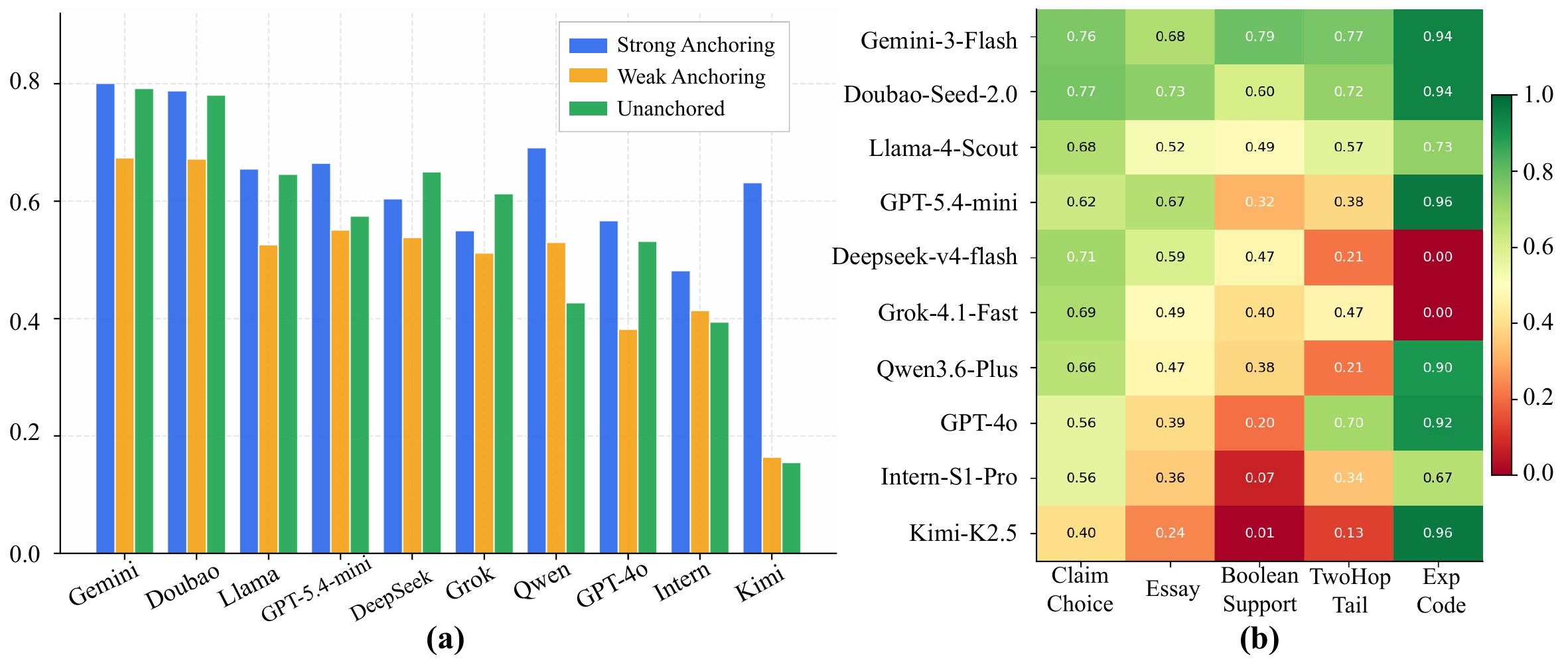}
    \caption{(a) Accuracy on Protein domain stratified by knowledge-graph
    anchoring level: \textit{Strong Anchoring}, \textit{Weak Anchoring}, and \textit{Unanchored}.
    Models sorted by overall accuracy. 
    (b) Accuracy by question type on Protein domain, rows sorted top-to-bottom by overall accuracy.}
    \label{fig:fig5}
    \vspace{-10pt}
\end{figure*}

\paragraph{The Trap of Partial Anchoring: When Incomplete Context Harms Reasoning.}
Figure~\ref{fig:fig5}(a) reveals a counter-intuitive pattern: accuracy does \emph{not} degrade monotonically as graph anchoring weakens.
Instead of a smooth downward gradient, six of the ten evaluated models—including \textsc{Gemini-3-Flash}, \textsc{Llama-4-Scout}, \textsc{GPT-5.4-mini}, and \textsc{GPT-4o}—exhibit a clear performance \emph{valley}, with \textit{Weak Anchoring} samples scoring lower than both \textit{Strong} and \textit{Unanchored} samples.
We attribute this to retrieval ambiguity: when supplied with partial graph context, an agent retrieves evidence that is informative but structurally incomplete, which encourages overconfident hallucinations grounded in fragmentary topology.
Completely unanchored questions, in contrast, force the model to fall back on internal parametric memory or broad search, bypassing the conflicting local structure altogether.
The collapse is most extreme in \textsc{Kimi-K2.5}, whose accuracy drops from $0.63$ (Strong) to $0.16$ (Weak)—a $0.47$-point cliff—while \textsc{Qwen3.6-Plus} shows the only strictly monotonic profile ($0.69 \!\to\! 0.53 \!\to\! 0.43$), suggesting it is the most graph-dependent model in the cohort.
This non-linear behavior indicates that, under our retrieval setup, partial topological context can be more harmful to autonomous agents than no anchoring at all.

\paragraph{Task Format Sensitivity: The Illusion of Aggregate Accuracy.}
Figure~\ref{fig:fig5}(b) shows that task format strongly modulates per-sample difficulty, exposing systemic weaknesses that aggregate scores conceal.
Within-model variance is striking: \textsc{Kimi-K2.5} alone spans nearly the entire accuracy scale, from $0.01$ on \texttt{boolean\_support} to $0.96$ on \texttt{experiment\_code}.
Across the cohort, models struggle markedly more with \texttt{boolean\_support} (avg.\ $0.375$) than with \texttt{claim\_choice} (avg.\ $0.645$), even though both are constrained, non-generative formats.
This gap reflects a specific weakness: \texttt{boolean\_support} forces a binary verdict on a single claim that requires deep evidence verification, penalizing models that terminate retrieval prematurely—\textsc{Intern-S1-Pro} illustrates this most clearly, scoring $0.07$ on Boolean tasks while remaining competitive on multiple-choice formats ($0.56$).
These format-induced disparities indicate that a single aggregate accuracy figure can mask substantial weaknesses in scientific evidence verification, making \emph{type-aware reporting} a necessary baseline.\footnote{Two models register $0.00$ on \texttt{experiment\_code} for their interaction trajectories that exceeded the maximum tool-call budget; this reflects resource consumption patterns rather than coding capability.} 
\section{Conclusion}
We present \textsc{Schema}, an evidence-grounded evaluation framework that reframes hallucination assessment in scientific agents from isolated query-level checks to topology-aware trajectory analysis.
\textsc{Schema} combines an automated concept-graph construction pipeline with two diagnostic instruments: a trajectory hallucination pipeline for large-scale claim-level severity measurement, and a multi-agent counterfactual module for fine-grained causal attribution on selected trajectories.
Together, these failures are structural rather than incidental.
Applied to two biomedical subdomains, \textsc{Schema} reveals that hallucinations concentrate at a small set of highly connected concept hubs, and that final-answer accuracy decouples from trajectory honesty—models often reach correct conclusions through structurally flawed reasoning.
For high-stakes scientific applications, terminal outcomes alone are insufficient.
\textsc{Schema} offers a foundation for the mechanism-level, topology-aware diagnostics that such systems demand.
Limitations are detailed in Appendix~\ref{app:Limitation}.

\section*{Acknowledgments and Disclosure of Funding}
This work was supported by the Shanghai Oriental Talents Program -- Top-notch Project (Grant No. BJZH2025127).

The authors declare no competing interests.

\clearpage

\bibliographystyle{unsrtnat}
\bibliography{content/my_bib}
\clearpage
\appendix
\section{Related Work}

\subsection{Scientific Agents}
The deployment of LLMs as scientific agents has rapidly evolved from isolated tool augmentation to end-to-end research automation~\citep{wang2024survey, xi2025rise}. 
Early efforts focused on tool-augmented domain experts, such as ChemCrow~\citep{m2024augmenting} and Coscientist~\citep{boiko2023autonomous}, which pair LLMs with software stacks or robotic platforms. 
This paradigm has recently expanded toward fully autonomous discovery. 
Systems like The AI Scientist~\citep{yamada2025ai} autonomously ideate, implement, and draft entire research pipelines. 
In the biomedical domain, efforts like BioDSA-1K~\citep{zhang2024benchmarking} and TruthHypo~\citep{xiong2025toward} evaluate data-science agents and hypothesis grounding. 
Yet, as this field shifts toward fully autonomous \emph{Agentic Science}~\citep{wei2025ai}, existing research overwhelmingly prioritizes \emph{capability expansion} over structural reliability. 
Current factuality assessments typically treat hallucination as a localized post-hoc filter, largely ignoring the agent's sequential reasoning trajectory.

\subsection{Agent Hallucination and Failure Attribution}
Traditional hallucination detection primarily audits stand-alone LLMs. 
Foundational works~\citep{min2023factscore, hu2024refchecker}) decompose generations into atomic claims, while single-turn benchmarks and graph-based metrics~\citep{li2023halueval}) probe factual breadth. 
However, these methods evaluate models in isolation, completely bypassing the interactive, multi-step trajectories characteristic of autonomous agents.
Consequently, an emerging thread targets \emph{agentic hallucination}. 
Benchmarks like AgentHallu~\citep{liu2026agenthallu} curate full interactive trajectories, while trajectory-aware metrics like TRACE~\citep{chen2026trace} highlight the ``high-score illusion''—where simple Pass@1 metrics conceal deeply flawed intermediate reasoning, an observation our experimental findings strongly corroborate.
Concurrently, the nascent field of \emph{failure attribution}~\citep{zhang2025agent}) transforms flat logs into causal graphs, applying counterfactual screening to distinguish root causes from propagated symptoms.
However, adapting these trajectory and attribution techniques to high-stakes scientific reasoning remains an unsolved challenge. While some efforts target biomedical text, none of the existing literature captures the deep conceptual interconnectivity of the domain.

\begin{table}[h]
\centering
\caption{\textsc{Schema} in the evolution of hallucination evaluations. 
Each row marks a step toward a more rigorous and structured evaluation. \cmark{} = supported, 
\xmark{} = not or partially supported.}
\label{tab:benchmark-comparison}
\footnotesize
\setlength{\tabcolsep}{6pt}
\renewcommand{\arraystretch}{1.05}
\begin{tabular}{l c c c c c}
\toprule
\multirow{2}{*}{\textbf{Evaluation}} 
& \textbf{Claim-Level} 
& \textbf{Trajectory} 
& \textbf{Knowledge} 
& \textbf{Topology-Based} 
& \textbf{Counterfactual} \\
& \textbf{Checking} 
& \textbf{Evaluation} 
& \textbf{Grounded}  
& \textbf{Weighting} 
& \textbf{Attribution} \\
\midrule
HalluHard      & \cmark & \cmark & \cmark & \xmark & \xmark \\
KGHaluBench    & \cmark & \xmark & \cmark & \cmark & \xmark \\
MIRAGE-Bench   & \xmark & \cmark & \xmark & \xmark & \xmark \\
DeepHalluBench & \cmark & \cmark & \cmark & \xmark & \xmark \\
\midrule
\rowcolor{citeblue!10}
\textbf{\textsc{Schema}} (ours) & \cmark & \cmark & \cmark & \cmark & \cmark \\
\bottomrule
\vspace{-10pt}
\end{tabular}
\end{table}
\section{Limitation}
\label{app:Limitation}

We discuss several limitations of the present work and corresponding directions for future research.
\paragraph{Domain Coverage.}
Although \textsc{Schema}'s graph-construction pipeline is domain-agnostic by design, our empirical evaluation is currently restricted to two biomedical subdomains—\textsc{Protein Domain} and \textsc{PathVQA-Enhanced}.
Scientific knowledge spans disciplines with widely varying ontological structures (e.g., chemistry, materials science, climate science), and the relative balance between concept hubs and long-tail rare concepts may differ substantially across fields.
Validating the generality of our topology-aware findings on additional disciplines remains an important next step.
\paragraph{LLM-as-Judge Reliance.}
The trajectory hallucination pipeline relies on an LLM judge to assign per-claim verdicts.
While we mitigate self-confirmation bias by using a separate model for extraction and judgment, restricting the judge to evidence-grounded entailment via system-prompt constraints, and caching all verdicts for reproducibility, the absence of large-scale expert-in-the-loop validation means that a small fraction of judge errors may persist.
Calibrating the pipeline against domain-expert annotations on a representative subset is a necessary direction for sharpening the reliability of automated severity scoring.
\paragraph{Cost of Counterfactual Attribution.}
The multi-agent counterfactual attribution module produces fine-grained causal localization but is substantially more expensive per trajectory than the trajectory pipeline, and is therefore applied only to a sampled subset of failures.
Scaling counterfactual analysis to full benchmarks would require either more efficient counterfactual construction or selective triggering policies; both are open problems we leave to future work.
\paragraph{Interaction Budget.}
All evaluated agents operate under a fixed maximum number of reasoning steps and tool calls.
Tasks that exceed this budget are recorded as failures, which can disadvantage models with less efficient tool-use strategies—two models in our cohort exhaust their budget on \texttt{experiment\_code} for this reason.
A more permissive budget would likely raise their scores at substantially higher inference cost; systematically characterizing this efficiency–capability trade-off is a meaningful follow-up.
\paragraph{Model Cohort.}
Our evaluation covers ten contemporary LLMs spanning open-source and closed-source families, but does not include fine-tuned scientific models or agents with custom retrieval pipelines.
Whether the patterns we report (hub concentration, accuracy-honesty decoupling) persist or sharpen under domain-specialized models is an open question that warrants dedicated study.

\section{Broader Impact}
\label{app:Boarder_Impact}

\textsc{Schema} is intended to support more rigorous and trustworthy evaluation of scientific agents, particularly in domains where errors in intermediate reasoning carry real downstream consequences.
By exposing how agents can produce correct final answers via structurally flawed reasoning trajectories, our work aims to discourage benchmarking practices that conflate terminal accuracy with epistemic reliability.
This is especially relevant for high-stakes scientific applications—drug discovery, clinical decision support, hypothesis generation—where a hallucinated mechanistic claim can cascade through downstream analyses long after the agent's surface output is accepted as correct.
At the same time, several aspects of our work warrant explicit consideration.
\paragraph{Risk of Benchmark Overfitting.}
Releasing a public benchmark inevitably creates incentives for models to be tuned toward its specific metric distribution rather than toward genuine improvements in scientific reasoning.
We mitigate this by providing two metrics ($\mathrm{HS}$ and $\mathrm{HS}^w$) that respond differently to error redistribution and by stratifying questions across five task formats with distinct reasoning bottlenecks—optimizing one metric or one format alone is unlikely to yield uniform gains.
We nonetheless encourage future users of \textsc{Schema} to report all metrics jointly and to treat any single number with appropriate caution.

\paragraph{LLM-as-Judge Dependence.}
Our trajectory hallucination pipeline relies on an LLM judge for per-claim verdicts.
While we constrain the judge to evidence-grounded entailment via system-prompt restrictions and use a separate model for extraction and judgment, large-scale automated judging can, in principle, propagate the preferences and blind spots of the underlying model family.
We view expert-in-the-loop calibration on a representative subset as the appropriate complement to fully automated evaluation, particularly when \textsc{Schema} is used to inform consequential decisions about model deployment.

\paragraph{Sensitive Domain Considerations.}
Both of our evaluation subsets draw from biomedical knowledge.
We deliberately frame \textsc{Schema} as a tool for diagnosing model behavior rather than as a clinical decision aid: a low hallucination severity score on \textsc{Schema} indicates that an agent passes a structured topology-aware audit, not that its outputs are safe to act on in patient care or laboratory settings.
Independent domain validation remains a prerequisite for any downstream deployment, and we caution against interpreting benchmark performance as a substitute for it.

\paragraph{Data Provenance.}
Our concept-graph construction draws on publicly accessible literature databases (e.g., PubMed, bioRxiv) and web-based scientific resources.
We use these sources strictly for evaluation-graph construction and citation tracking, and the released benchmark contains derived knowledge triples and question artifacts rather than redistributed full-text content, in line with standard fair-use practice for evaluation research.

\paragraph{Positive Use Cases.}
Beyond benchmarking, the topology-aware framework underlying \textsc{Schema} can support targeted improvement of scientific agents: identifying which concept hubs concentrate failures, which question formats expose a model's specific reasoning bottlenecks, and which trajectory steps a counterfactual analysis attributes errors to.
We hope these signals motivate more diagnostic—rather than purely competitive—evaluation practices in the scientific-agent community.


\section{Detailed Evaluation Experimental Setup}
\label{app:experimental_setup}

\paragraph{Baselines.}
The full cohort of 10 evaluated models includes \texttt{gpt-4o}~\citep{hurst2024gpt}, \texttt{gpt-5.4-mini}~\citep{singh2025openai}, \texttt{gemini-3-flash-preview-thinking}, \texttt{grok-4-1-fast-reasoning}, \texttt{kimi-k2.5}~\citep{team2026kimi}, \texttt{qwen3.6-plus}, \texttt{deepseek-v4-flash}, \texttt{llama-4-scout}~\citep{adcock2026llama}, \texttt{doubao-seed-2.0-pro} and \texttt{intern-s1-pro}~\citep{zou2026intern}. 
Sampling temperature was fixed to 0 or otherwise to the lowest provider-supported value, with top\_p set to 1. 

\paragraph{Workflow Configuration.}
The workflow equips the \texttt{EvalAgent} with four tools: \texttt{literature\_search}, \texttt{literature\_fetch}, \texttt{web\_search}, and \texttt{web\_fetch}.
Literature tools are backed by our PubMed-graph pipeline, while web tools are powered by Bright Data.
Each dialogue allows a maximum of 12 turns and is constrained by a 600-second wall-clock budget.
To prevent data leakage during hallucination evaluation, tool execution results are explicitly elided from the trajectory logs, ensuring that retrieved evidence does not contaminate the claim extraction process. 


\paragraph{Cost.} The full evaluation across 10 LLM agents and two biomedical subdomains incurred approximately \$350 USD in total API and infrastructure costs.

\paragraph{Tool Infrastructure.} Web search and literature retrieval are routed through Bright Data SERP. 
Agent orchestration and trajectory analysis run on commodity CPU/memory hardware.

\section{Detailed Benchmark Construction Pipeline}
\subsection{Concept-Graph Construction}
\label{app:concept_graph}

\paragraph{Cascaded Term Extraction.}
Candidate biomedical seed terms are extracted via a two-stage cascade.
A rule-based parser first captures terms matching curated domain patterns (e.g., gene/protein nomenclature, MeSH-style entity formats); only residual text not covered by the parser is forwarded to an LLM-based extractor.
This design contains LLM cost while preserving recall on irregular surface forms.

\paragraph{Retrieval Scoring.}
Given a seed $\sigma$, retrieved papers $p$ from PubMed, bioRxiv, and related literature corpora are ranked by a weighted combination of SapBERT~\citep{liu2021self} semantic similarity and a normalized Journal Impact Factor:
\begin{equation}
\mathrm{score}(p) = w_{\text{sem}} \cdot \cos \bigl(\phi_{\text{SapBERT}}(\sigma),\, \phi_{\text{SapBERT}}(p)\bigr) + w_{\text{imp}} \cdot \min \bigl(\mathrm{JIF}(p)/100,\, 1\bigr).
\label{eq:retrieval-score}
\end{equation}
Papers exceeding $\theta_{\text{score}}$ are cached for downstream extraction.
The semantic term dominates ($w_{\text{sem}} \gg w_{\text{imp}}$ in our setting); the impact term acts only as a soft tie-breaker among semantically comparable candidates.

\paragraph{Triple Extraction with Confidence.}
For each cached text, an LLM extracts knowledge triples together with a self-reported confidence score.
We retain confidence as a per-edge attribute rather than applying it as a hard filter, enabling downstream analyses to trace edge provenance and quality.

\paragraph{Node Fusion.}
Node pairs are fused when their BGE-Large~\citep{luo2024bge} embedding similarity exceeds $\theta_f$ \emph{and} they share the same entity type under $\eta$.
The type-consistency constraint is critical: it prevents semantically close but ontologically distinct entities (e.g., a protein and its encoding gene) from being collapsed, which would otherwise corrupt the downstream typed reasoning.

\paragraph{Topology-Aware Subgraph Sampling.}
To exploit the conceptual interconnectivity of $\mathcal{G}$, we extract two classes of subgraphs as question substrates. 
\emph{Single-hop subgraphs} $(head, relation, tail)$ are sampled with a uniqueness key to control duplication. 
\emph{Two-hop subgraphs} chain compatible single hops ($A\!\rightarrow\!B\!\rightarrow\!C$) where the bridging node $B$ is graph-unique, explicitly facilitating multi-hop reasoning questions. For \texttt{experiment\_code} tasks, the sampler emits candidates across three difficulty tiers (easy/medium/hard), utilizing difficulty suffixes to ensure all variants survive deduplication.

\paragraph{Evidence Profiling and Gating.}
To ensure factual fidelity, an evidence profiler evaluates the extracted triples, computing a combined evidence strength ($\in \{\textit{weak}, \textit{medium}, \textit{strong}\}$) based on relation priors and lexical hedge density within the text chunks. This metric serves a critical dual purpose. First, it \textbf{gates legal question types}: while \texttt{claim\_choice} and \texttt{essay} accept any strength, \texttt{boolean\_support} demands at least \emph{medium}, and \texttt{two\_hop\_tail} strictly requires \emph{strong} evidence on both hops. (Note that \texttt{experiment\_code} bypasses this textual gate as it relies on programmatic sandbox execution). Second, it \textbf{drives claim wording}: the phrasing is strictly anchored to the evidence strength (e.g., \emph{``suggests a reported association''} vs. \emph{``supports that $X$ regulates $Y$''}), preventing the benchmark from over-claiming.

\paragraph{Template and Programmatic Generation.}
For text-based question types (\texttt{claim\_choice}, \texttt{boolean\_support}, \texttt{two\_hop\_tail}, \texttt{essay}), questions are synthesized by merging per-type templates with the strength-calibrated claims. Crucially, distractors for multiple-choice formats are systematically drawn from same-type entities within $\mathcal{T}$ sharing the same graph context; samples failing to yield at least two viable distractors are silently dropped.

For the \texttt{experiment\_code} type, we employ a \emph{spec-then-verify} programmatic strategy. An LLM is prompted to synthesize a comprehensive specification including data code, main orchestrations, incomplete functions, and unit tests. Difficulty is enforced via deterministic blank selection: \emph{easy} blanks a single helper, \emph{medium} blanks all helpers, and \emph{hard} additionally blanks the orchestration logic. If the LLM spec fails sandbox validation (detailed below), the error trace is fed back for iterative refinement (up to three retries). Exhausted LLM paths gracefully fall back to four hardcoded blueprints (e.g., \textit{biomarker\_screening}, \textit{dose\_response}).

\paragraph{Multi-Stage Validation Stack.}
Every generated sample must pass a rigorous, parallelized validation stack. Failure at any stage triggers immediate rejection.

\textit{(i) Rule-Based Guardrails.} We enforce strict topological and lexical constraints: (1) \textbf{Evidence support}: The triple's evidence string must exhibit $\geq 80\%$ token overlap with the cited chunk. (2) \textbf{Minimum double-check}: Claims must be supported by at least two distinct documents/chunks. (3) \textbf{No answer leakage}: Stems are scrubbed of title-prefixes to prevent trivial entity matching. (4) Answer uniqueness, type-evidence compatibility, and metadata completeness.

\textit{(ii) Sandbox Execution Gate.} For \texttt{experiment\_code}, the synthesized codes and unit tests are bundled and executed in an isolated Python sandbox via \texttt{sandbox\_fusion} (protected by TCP probes and a 20-second timeout). Acceptance strictly requires the \emph{reference} solution to pass all tests while the \emph{blanked} solution fails at least one, mathematically guaranteeing non-triviality. Soft infrastructure failures gracefully degrade to rule-only acceptance.

\textit{(iii) LLM-as-Judge Cross-Validation.} For text questions, a judge evaluates the claim against an evidence bundle composed of local chunks and live PubMed queries (top-$k=3$). For \texttt{experiment\_code}, a \emph{second-opinion judge} reviews the code against the original graph claim to detect off-topic or direction-flipped logic (e.g., \emph{inhibits} $\leftrightarrow$ \emph{activates}) that unit tests miss. Verdicts map to \textsc{supported} (accept), \textsc{contradicted}/\textsc{insufficient} (reject), or \textsc{model\_unavailable} (degrade to rule-only). All judge inferences are content-hashed and globally cached to enable rapid, reproducible re-runs.

\paragraph{Deduplication and Export.}
Following validation, questions sharing identical text are collapsed to preserve distractor ordering integrity. The exporter compiles the final corpus alongside a comprehensive diagnostic summary (evidence distributions, rejection typologies) to ensure full benchmark reproducibility and auditability.

\subsection{Question Generation}
\label{app:question_generation}

\begin{table}[h]
\centering
\caption{Overview of the five question types in \textsc{Schema}. Each type isolates a distinct cognitive operation so that failures map cleanly to specific reasoning bottlenecks.}
\label{tab:qtype-overview}
\footnotesize
\setlength{\tabcolsep}{5pt}
\renewcommand{\arraystretch}{1.2}
\begin{tabular}{l l l c}
\toprule
\textbf{Type} & \textbf{Format} & \textbf{Capability Probed} & \textbf{Evidence Gate} \\
\midrule
\textsc{ClaimChoice}    & 4-way claim selection      & Discriminative mapping       & Any           \\
\textsc{BooleanSupport} & Binary support judgment    & Calibration                  & \textsc{Medium}+ \\
\textsc{TwoHopTail}     & Bridging tail prediction   & Multi-hop factual recall     & \textsc{Strong} \\
\textsc{Essay}          & Open-ended explanation     & Synthesis \& faithfulness    & Any           \\
\textsc{ExperimentCode} & Executable Python + tests  & Operationalization           & Sandbox-gated  \\
\bottomrule
\end{tabular}
\end{table}

Building upon the hierarchical knowledge graph $\mathcal{G}$ established in Section~\ref{sec:method-graph-construction}, we design our generation pipeline to translate these domain priors into an \emph{evidence-grounded} benchmark.
Most KG-derived benchmarks generate questions directly from relation labels. We instead tie each question's wording, eligibility, and validation to the supporting sentence in the source paper.
The generator directly consumes the assembled multigraph $\mathcal{G}$ and the globally cached text chunks from the retrieval phase.

\paragraph{Question Type Inventory.}
The five question types in \textsc{Schema} are designed to be \emph{orthogonal}: each isolates one cognitive operation, so that an agent's failure on a given item maps cleanly to a specific reasoning bottleneck.
Table~\ref{tab:qtype-overview} summarizes their formats, the capabilities they target, and the evidence-strength gating each requires.

\textsc{ClaimChoice} presents four candidate claims drawn from the same neighborhood of $\mathcal{G}$ and asks the agent to identify which one is most directly supported by the evidence; distractors are entities of the same type sharing graph context, so the task isolates \emph{discriminative mapping}—telling apart adjacent but inequivalent claims.
\textsc{BooleanSupport} presents a single claim and asks for a binary supported / not-supported judgment, which probes \emph{calibration}: whether the agent over-claims or refuses in the absence of comparative anchors. Because binary judgment on ambiguous claims is intrinsically unfair, this type is gated to subgraphs with at least \textsc{medium} evidence.
\textsc{TwoHopTail} fills the bridging tail in a chain $A \to B \to C$, evaluating \emph{multi-hop factual recall}; both hops must independently meet \textsc{strong} evidence to admit the question.
\textsc{Essay} requests an open-ended explanation paired with a strength-calibrated reference answer, evaluating \emph{synthesis and natural-language faithfulness}; downstream scoring is left to benchmark users (LLM-as-Judge, ROUGE, or human evaluation), and we provide only the grounded reference.
\textsc{ExperimentCode} requires translating a biological claim into executable Python code with unit tests, evaluating \emph{operationalization}; correctness is enforced programmatically through sandbox execution rather than evidence-strength gating.

\paragraph{Evidence Profile Computation.}
The evidence profile $s$ is computed from three independent signals per subgraph and aggregated by a deterministic decision rule.

\textit{(1) Relation strength prior.} Each relation $r \in \mathcal{R}$ is mapped to a tier through an ontology lookup with a hardcoded fallback. Causal/directional relations (\textit{e.g.}, \textit{activates}, \textit{inhibits}, \textit{upregulates}, \textit{downregulates}, \textit{promotes}) are tagged \textsc{strong}; weak associative relations (\textit{associated\_with}, \textit{part\_of}) are tagged \textsc{weak}; remaining relations default to \textsc{medium}. For multi-hop subgraphs, we take the weakest tier across hops.

\textit{(2) Hedge density.} We score the lexical hedge density of supporting text by counting occurrences of hedge markers (\textit{suggest, may, might, possibly, associated, correlat, linked, \ldots}) and normalizing by a saturation cap of three: $\text{hedge}(\cdot) = \min(\text{count}/3, 1)$. We additionally compute the hedge score over the full source chunk and take the per-hop maximum, ensuring that strong-looking evidence inside a hedged context is not falsely promoted.

\textit{(3) Independent support.} For each hop, we count the number of distinct documents/chunks supporting the underlying triple after deduplication; the multi-hop aggregate is the minimum across hops.

The three signals are combined by an asymmetric AND/OR rule: a subgraph is labeled \textsc{strong} \emph{only if} all hops are \textsc{strong}, support count $\geq 2$, and hedge $< 0.34$; it is labeled \textsc{weak} if any hop is weak, hedge $\geq 0.67$, or any hop has zero independent support; otherwise it is \textsc{medium}. The asymmetry is deliberate: it is acceptable to under-state strength (yielding more conservative phrasing), but never to over-state it, since types like \textsc{TwoHopTail} would otherwise admit weakly grounded items.

\paragraph{Validation Configuration.}
Rule-based guardrails operate on different signals depending on type: lexical-overlap and exact-match constraints apply to selection-style types (\textsc{ClaimChoice}, \textsc{BooleanSupport}, \textsc{TwoHopTail}), whereas \textsc{Essay} and \textsc{ExperimentCode} are evaluated by their own type-appropriate gates (judge-based and sandbox-based, respectively).
For \textsc{ExperimentCode}, the masked variant is produced by an AST-aware editor that replaces the body of pre-declared helper functions with \texttt{pass} while preserving signatures, docstrings, imports, and synthetic data generation. Difficulty controls which functions are blanked: \textsc{easy} blanks one helper, \textsc{medium} blanks all listed helpers, \textsc{hard} additionally blanks the orchestration logic. Sandbox acceptance requires that the reference solution pass all unit tests and the masked variant fail at least one; failure in compilation also counts as a meaningful blank.
For LLM-as-Judge cross-validation, the judge is restricted by system prompt to use only the supplied evidence bundle (local chunks plus top-$k$ retrieved passages) and to refrain from invoking outside knowledge, framing the task as \emph{evidence-grounded entailment} rather than free-form factuality. Judge outputs (verdict, score, rationale, issue tags) are content-hashed and cached for reproducibility; failures of the judge service degrade gracefully to rule-only acceptance with the corresponding mode flag recorded.
Full prompt designs are deferred to Prompt~\ref{box:cross_valid}.


\paragraph{Subgraph Sampling.} To exploit the inherent interconnectivity of $\mathcal{G}$, we first sample single- or two-hop subgraphs for each candidate triple $(h, r, t)$ and recover their cited chunks from the retrieval cache.
Rather than treating edges as equally informative, we compute an \emph{evidence profile} for the chunk.
This profile combines a relation-strength prior with the lexical hedge density of the cited sentence, yielding a discrete strength label $s \in \{\textsc{weak}, \textsc{medium}, \textsc{strong}\}$.
Crucially, the label serves a dual purpose: it \emph{gates} which question types are admissible (e.g., \textsc{TwoHopTail} strictly requires both hops to be strong) and \emph{modulates} the surface form of the generated claim, ensuring that strong evidence yields assertive wording while weak evidence enforces hedged phrasing.


\subsection{Hallucination Pipeline Implementation Details.}
\label{app:hallu_pipeline}
The hallucination pipeline operates on the per-sample agent trajectories emitted by the evaluation runner. 
To focus on structural reasoning failures, the pipeline defaults to an error-targeted mode (\texttt{is\_correct=false}), computing severity exclusively on trajectories that yield incorrect terminal answers.
The pipeline executes through the following deterministic stages:
\textbf{1. Preprocessing and Leakage Prevention.} For each agent message, we strictly filter out the bodies of \texttt{ToolCallExecutionEvent}s. 
Only the agent's natural-language reasoning and the arguments of \texttt{ToolCallRequestEvent}s are retained. This isolation ensures that raw retrieved content cannot leak into the claim extractor and skew the evaluation.
\textbf{2. Claim Extraction and Normalization.} A designated extractor (\texttt{gpt-4o}, $T{=}0$, JSON mode) decomposes the filtered steps into atomic factual claims. 
Concept strings are canonicalized via \texttt{pubmed\_graph.normalize}, utilizing a BGE-Large embedding fallback with a cosine similarity threshold of $0.85$. 
Claims sharing a canonical concept are batched into a single bucket, ensuring the judge evaluates the model's complete, unified stance on any given entity.
\textbf{3. Evidence Grounding.} For each concept bucket, evidence is gathered via a strict, short-circuiting priority chain that terminates at the first non-empty layer: (1) gold \texttt{supporting\_chunks} $\rightarrow$ (2) \texttt{global\_graph} 1-hop neighborhood (retrieved via BGE node embeddings cached in \texttt{bge\_nodes.npy}) $\rightarrow$ (3) \texttt{web\_search} $\rightarrow$ (4) \texttt{literature\_search}.
\textbf{4. Deterministic Judgement.} The judge LLM processes each bucket via a single JSON-mode call, assigning discrete qualitative verdicts: \texttt{supported}, \texttt{refuted}, or \texttt{unverifiable}. 
Crucially, a deterministic Python script maps these verdicts to strict numerical severity penalties ($0.0$, $1.0$, and $0.5$, respectively) before aggregation, systematically preventing the LLM from generating arbitrary numeric scores. 
\textbf{5. Judge Allocation and Caching.} To prevent self-evaluation bias, \texttt{gpt-4o} serves as the default judge, with the sole exception that \texttt{intern-s1-pro} is substituted when grading \texttt{gpt-4o}'s own trajectories. 
Finally, both extractions and judgements are heavily disk-cached (keyed by step-text hash and concept-evidence tuple hashes, respectively), enabling computationally free re-runs for identical configurations.

\subsection{Mechanism-Attributed Evaluation Framework}
\label{app:metrics}

Let the evaluation set be $\mathcal{D} = \{s_i\}_{i=1}^N$, where each sample $s_i = (q_i, a_i^\star, \mathcal{O}_i, \mathcal{G}_i, w_i, \mathrm{type}(s_i))$ contains a question, gold answer, options, graph-grounded context, benchmark weight $w_i > 0$, and question type. For a candidate agent $\pi_c$, we evaluate both its final answer $\hat{a}_{m, i}$ and its complete reasoning trajectory $\tau_{m, i}$.

\paragraph{Accuracy Measurement.}
\label{sec:metrics-acc}
The per-sample correctness $c_{m, i} \in [0, 1]$ adapts to the question type: boolean logic for objective QA (\textit{claim\_choice}, \textit{boolean\_support}, \textit{two\_hop\_tail}, \textit{vqa}), GPT-4o-based judging for \textit{essay}, and unit-test pass rates for \textit{experiment\_code}. We report both unweighted ($\mathrm{Acc}$) and benchmark-weighted ($\mathrm{WAcc}$) accuracy:
\begin{equation}
\mathrm{Acc}(m) = \frac{1}{N} \sum_{i=1}^N c_{m,i}, \qquad \mathrm{WAcc}(m) = \frac{\sum_{i=1}^N w_i \, c_{m,i}}{\sum_{i=1}^N w_i}.
\end{equation}
Crucially, $w_i$ encapsulates the graph tier and evidence strength of the sample, systematically down-weighting answers reliant on weak supporting evidence.

\paragraph{Aggregation Formulas.}
For trajectory $\tau_{m,i}$ with claims $\{c_1, \ldots, c_n\}$, per-claim severity scores $\rho(c_j) \in \{0, 0.5, 1\}$, and graph weights $w_g(c_j)$ (Eq.~\ref{eq:wg}), the unweighted and weighted hallucination severities are:
\begin{equation}
\mathrm{HS}(\tau) = \frac{1}{n}\sum_{j=1}^{n} \rho(c_j), \quad
\mathrm{HS}^w(\tau) = \frac{\sum_{j=1}^{n} w_g(c_j) \cdot \rho(c_j)}{\sum_{j=1}^{n} w_g(c_j)}.
\end{equation}
We report both metrics in all experimental tables. The unweighted form preserves comparability with prior hallucination benchmarks, while the weighted form penalizes errors at high-connectivity concepts more heavily, reflecting their downstream blast radius.

\paragraph{Claim-to-Node Mapping $\eta$.}
The map $\eta$ first attempts a normalized lexical match of the claim string against the union of node identifiers and node aliases extracted from the graph (returning $\cos = 1.0$ on a hit). On a miss, it embeds the claim via BGE-Large and returns the node maximizing cosine similarity over precomputed node-label embeddings, gated by a floor $\theta_{\cos} = 0.6$. Claims with no node above the floor are treated as off-graph and receive baseline weight $w_g = 1$.

\paragraph{Limitations of the Weighting Scheme.}
Off-graph claims (those failing $\eta$ at $\theta_{\cos}$) receive $w_g = 1$ rather than $w_g = 0$; they remain in both numerator and denominator of $\mathrm{HS}^w$, but at a discount relative to in-graph hub claims. We deliberately retain both $\mathrm{HS}$ and $\mathrm{HS}^w$ as headline metrics: an adversarial strategy of fabricating exclusively off-graph concepts cannot reduce $\mathrm{HS}^w$ without simultaneously inflating $\mathrm{HS}$. Combined with a permissive cosine floor $\theta_{\cos} = 0.6$ that admits paraphrases and aliases, the effective space of "off-graph but plausible" concepts is small.


\paragraph{Topology-Aware Severity Weighting.}
\label{sec:metrics-topology-weight}
Standard hallucination metrics treat all concepts equally, ignoring the intrinsic interconnectivity of scientific domains. However, a hallucination on a foundational hub (e.g., a central protein) triggers severe cascading errors, whereas an error on a peripheral node remains localized. To capture this structural asymmetry, we explicitly penalize errors on load-bearing entities using a graph-degree-based weight:
\begin{equation}
w_g(c) =
\begin{cases}
    1 + \log\bigl(1 + \deg_{\mathcal{K}}(\eta(c))\bigr) & \text{if } \eta(c) \text{ exists and } \cos \geq \theta_{\cos}, \\
    1 & \text{otherwise,}
\end{cases}
\label{eq:hs-weight}
\end{equation}
where $\deg_{\mathcal{K}}$ denotes the undirected node degree in the global knowledge graph $\mathcal{K}$. The logarithmic scaling prevents hyper-connected hubs from disproportionately dominating the metric. 

Let $\mathcal{C}_{m,i}$ denote the multiset of all judged claims for sample $i$. We formulate the \textbf{Topology-Weighted Hallucination Severity ($\mathrm{HS}^w$)} as:
\begin{equation}
\mathrm{HS}^{w}_{m,i} = \frac{\sum_{k \in \mathcal{C}_{m,i}} w_g\bigl(\tilde{e}(k)\bigr) \cdot \rho\bigl(v(k)\bigr)}
      {\sum_{k \in \mathcal{C}_{m,i}} w_g\bigl(\tilde{e}(k)\bigr)}.
\label{eq:hs-weighted}
\end{equation}
Alongside $\mathrm{HS}^w$, we also compute the unweighted Hallucination Rate ($\mathrm{HR}_{m,i}$) and the binary Hallucination Flag ($\mathrm{HF}_{m,i} = \mathbb{1}[\exists v(k) = \textsc{refuted}]$) to capture the raw error frequency.

\paragraph{Aggregation and Implementation.}
\label{sec:metrics-impl}
Metrics are aggregated at both the \emph{macro} (per-sample mean) and \emph{micro} (claim-weighted) levels across all candidates $\pi_c$.
To provide granular insights, performance is further sliced by question type, graph tier (Core/Peripheral), and node type (e.g., \textit{Protein}, \textit{Disease}).

\section{Protein Domain Question-Type Examples}
\label{app:bench_a_examples}

Bench~A (\texttt{paired\_enhanced\_v2\_balanced}) contains five question types: \texttt{claim\_choice}, \texttt{two\_hop\_tail}, \texttt{boolean\_support}, \texttt{essay}, and \texttt{experiment\_code}. We display one representative, evidence-verified sample per type. For each multiple-choice item, the correct option is marked with a $\bigstar$; the supporting passage shown below the question is the verbatim chunk from the source paper that justifies the labelled answer.

\bigskip

\begin{tcolorbox}[
    breakable,
    colframe=black!75!black,
    colback=gray!10!white,
    colbacktitle=gray!30!white,
    title=\textbf{Example A.1 --- \texttt{claim\_choice}\quad(\texttt{qg\_000048})},
    coltitle=black,
    boxrule=0.3mm,
    rounded corners,
    top=6pt, bottom=6pt, left=6pt, right=6pt,
    fonttitle=\bfseries,
    before upper={\parindent15pt}
]

\textbf{Source paper:} \textit{Cancer cachexia: molecular mechanisms and treatment strategies} (PMID:37217930).

\vspace{0.5em}
\textbf{Question.} Based on the reported evidence from the paper above, which candidate claim is most directly supported by the provided evidence?

\vspace{0.5em}
\textbf{Options.}

\hspace{10pt}\hspace{1em}A. The evidence supports a contextual relationship in which AMPK$\alpha$1 activates PHD2.

\hspace{10pt}\hspace{1em}B. The evidence supports a contextual relationship in which GST--hnRNP E1 activates the translational signal.

\hspace{10pt}\hspace{1em}C. The evidence supports a contextual relationship in which Heterogeneous nuclear ribonucleoprotein E1 activates p53 and p21.

\hspace{10pt}\hspace{1em}$\bigstar$ D. The evidence supports a contextual relationship in which Cytokines activate NF-$\kappa$B.

\vspace{0.5em}
\textbf{Supporting passage.} ``\textit{Cytokines from tumors and immune cells induce activation of transcription factor NF-$\kappa$B, leading to UPS and ALS activation, which leads to muscle wasting.}''

\vspace{0.5em}
\textbf{Underlying triple.} \texttt{(Cytokines, activates, NF-$\kappa$B)} with confidence $0.95$.

\end{tcolorbox}

\bigskip

\begin{tcolorbox}[
    breakable,
    colframe=black!75!black,
    colback=gray!10!white,
    colbacktitle=gray!30!white,
    title=\textbf{Example A.2 --- \texttt{two\_hop\_tail}\quad(\texttt{qg\_000802\_\_d2})},
    coltitle=black,
    boxrule=0.3mm,
    rounded corners,
    top=6pt, bottom=6pt, left=6pt, right=6pt,
    fonttitle=\bfseries,
    before upper={\parindent15pt}
]

\textbf{Source paper:} \textit{Peripheral Vascular Calcification} (PMID:41537264).

\vspace{0.5em}
\textbf{Question.} Based on the reported evidence from the paper above, which intermediate pathway is best supported by the evidence chain in which Niclosamide inhibits an intermediate entity, and that intermediate entity participates in peripheral vascular calcification?

\vspace{0.5em}
\textbf{Options.}

\hspace{10pt}\hspace{1em}A. AKT/mTOR signaling pathway

\hspace{10pt}\hspace{1em}$\bigstar$ B. Wnt signaling pathway

\hspace{10pt}\hspace{1em}C. AKT signaling pathways

\hspace{10pt}\hspace{1em}D. AKT signaling pathway

\vspace{0.5em}
\textbf{Supporting passage.} ``\textit{Inhibiting Wnt signaling, exemplified by the use of Niclosamide, has been shown to effectively suppress the expression of osteogenic genes, including Runx2, and reduces vascular calcification.}''

\vspace{0.5em}
\textbf{Underlying triple chain.} \texttt{(Niclosamide, inhibits, Wnt signaling pathway)} $\rightarrow$ \texttt{(Wnt signaling pathway, reduces, vascular calcification)}; both hops are visible in the single sentence above.

\end{tcolorbox}

\bigskip

\bigskip

\begin{tcolorbox}[
    breakable,
    colframe=black!75!black,
    colback=gray!10!white,
    colbacktitle=gray!30!white,
    title=\textbf{Example A.3 --- \texttt{boolean\_support}\quad(\texttt{qg\_000999})},
    coltitle=black,
    boxrule=0.3mm,
    rounded corners,
    top=6pt, bottom=6pt, left=6pt, right=6pt,
    fonttitle=\bfseries,
    before upper={\parindent15pt}
]

\textbf{Source paper:} \textit{CYLD Limits Neutrophil-Driven Psoriatic Inflammation} (\textit{Inflammation} 2026, \texttt{10.1007/s10753-026-02452-3}).

\vspace{0.5em}
\textbf{Question.} Based on the reported evidence from the paper above, is the following scientific claim supported by the provided evidence set: \textit{The reported evidence suggests that CYLD is upregulated in psoriasis}?

\vspace{0.5em}
\textbf{Options.}

\hspace{10pt}\hspace{1em}$\bigstar$ A. Supported

\hspace{10pt}\hspace{1em}B. Not supported

\vspace{0.5em}
\textbf{Supporting passage.} ``\textit{The results showed that CYLD expression was significantly upregulated in lesional skin of psoriasis patients; CYLD$^{-/-}$ mice displayed more severe psoriasiform symptoms.}''

\vspace{0.5em}
\textbf{Underlying triple.} \texttt{(CYLD, upregulated\_in, Psoriasis)} with confidence $0.89$.

\end{tcolorbox}

\bigskip

\begin{tcolorbox}[
    breakable,
    colframe=black!75!black,
    colback=gray!10!white,
    colbacktitle=gray!30!white,
    title=\textbf{Example A.4 --- \texttt{essay}\quad(\texttt{qg\_000131})},
    coltitle=black,
    boxrule=0.3mm,
    rounded corners,
    top=6pt, bottom=6pt, left=6pt, right=6pt,
    fonttitle=\bfseries,
    before upper={\parindent15pt}
]

\textbf{Source paper:} \textit{Infection-Associated Thymic Atrophy} (PMID:34113341).

\vspace{0.5em}
\textbf{Question.} Based on the reported evidence from the paper above, explain the relationship between Vpu and the release of virion, and discuss the strength of the supporting findings.

\vspace{0.5em}
\textbf{Reference answer.} The reported evidence suggests that Vpu enhances the release of virion. Supporting evidence: \textit{The Vpu protein of HIV-1 enhances the release of virion from infected cells.}

\vspace{0.5em}
\textbf{Supporting passage.} ``\textit{The Vpu protein of HIV-1 enhances the release of virion from infected cells. However, the Simian--Human Immunodeficiency Virus with the deletion of Vpu sequence (novpuSHIV KU-1bMC33) can still result in severe thymic atrophy in macaques.}''

\vspace{0.5em}
\textbf{Underlying triple.} \texttt{(Vpu, enhances, release of virion)}.

\end{tcolorbox}

\bigskip

\begin{tcolorbox}[
    breakable,
    colframe=black!75!black,
    colback=gray!10!white,
    colbacktitle=gray!30!white,
    title=\textbf{Example A.5 --- \texttt{experiment\_code}\quad(\texttt{qg\_999001})},
    coltitle=black,
    boxrule=0.3mm,
    rounded corners,
    top=6pt, bottom=6pt, left=6pt, right=6pt,
    fonttitle=\bfseries,
    before upper={\parindent15pt}
]

\textbf{Scientific claim.} \textit{Insulin-like growth factor 2 upregulates Myogenic differentiation 1.}

\vspace{0.5em}
\textbf{Research direction.} IGF-II upregulates MyoD expression in myoblast differentiation. Compute the Pearson correlation between IGF-II and MyoD expression levels and return a support flag.

\vspace{0.5em}
\textbf{Evidence summary.} ``\textit{IGF-II plays a direct role in the differentiation of mesoderm into myoblasts by upregulating the expression of MyoD, a key factor determining musculoskeletal fate.}'' (PMID:35741015)

\vspace{0.5em}
\textbf{Provided synthetic data} (\texttt{data\_en.py}, deterministic positive-correlation fixture):

\hspace{10pt}\texttt{IGF2 = [1.0, 2.0, 3.0, 4.0, 5.0]}

\hspace{10pt}\texttt{MyoD = [2.0, 4.0, 6.0, 8.0, 10.0]}

\vspace{0.5em}
\textbf{Skeleton to complete} (\texttt{main\_en.py}):

\hspace{10pt}\texttt{def compute\_correlation(data):}

\hspace{20pt}\texttt{\#\# Pearson correlation between data['IGF2'] and data['MyoD']}

\hspace{20pt}\texttt{pass  \#\# [Please complete the code]}

\vspace{0.5em}
\hspace{10pt}\texttt{def summarize\_igf2\_myo\_upregulation():}

\hspace{20pt}\texttt{data = load\_igf2\_myo\_data()}

\hspace{20pt}\texttt{corr = compute\_correlation(data)}

\hspace{20pt}\texttt{return \{'correlation': corr, 'supported': corr > 0\}}

\vspace{0.5em}
\textbf{Reference solution.} A standard Pearson correlation: centre both vectors, divide their dot product by the product of their L2 norms; under the supplied fixture this returns $\rho = 1.0$ and \texttt{supported} = \texttt{True}, consistent with the evidence sentence.

\end{tcolorbox}

\section{Prompts}

\begin{tcolorbox}[
    breakable,
    colframe=black!75!black,
    colback=gray!10!white,
    label=box:answer_prompt,
    colbacktitle=gray!30!white,
    title=\textbf{Agent $\pi_s$ System Prompt},
    coltitle=black,
    boxrule=0.3mm,
    rounded corners,
    top=6pt, bottom=6pt, left=6pt, right=6pt,
    fonttitle=\bfseries,
    before upper={\parindent15pt}
]
\begin{Verbatim}[breaklines=true,breakanywhere=true,fontsize=\small]
## General Policy

1. **Actively use ToolUniverse tools** when they can help answer scientific questions.
2. For scientific/technical questions, always attempt to find relevant tools before relying solely on general knowledge.
3. Use `find_tools` to discover appropriate ToolUniverse tools for the task.
4. Balance tool usage with reasoning - tools should enhance, not replace, analytical thinking.

## Tool Search Strategy

1. **When to search for tools**:
   - Scientific calculations, simulations, or data analysis
   - Domain-specific queries (biology, chemistry, physics, etc.)
   - Questions requiring specialized knowledge or databases
   - Tasks that could benefit from computational tools

2. **How to search**:
   - Build focused queries from key entities and domain terms
   - Example: "protein structure analysis PDB", "molecular dynamics simulation", "gene expression analysis"
   - If first search doesn't find suitable tools, try a broader query
   - Maximum 2-3 search attempts per question to avoid loops

3. **Tool validation**:
   - Check if the tool directly addresses the question
   - Verify required parameters are available
   - Ensure the tool output will help answer the question
   - Always use the **exact parameter names** defined in each tool's schema
   - If unsuitable, proceed with reasoning and available knowledge

## General Web Retrieval Strategy

1. **When to use general web retrieval**:
   - Questions requiring up-to-date information not covered by ToolUniverse tools
   - Verifying facts, finding references, or retrieving specific data from websites
   - When ToolUniverse tools are not available or suitable for the task

2. **How to use general web retrieval**:
   - Call `web_search` first to get an overview of available sources (snippets + URLs)
   - If the snippets contain enough information, use them directly - no further fetching needed
   - If more detail is required, call `web_fetch` with 1-3 URLs from the search results
   - `web_fetch` is for full text retrieval from known URLs and automatically handles both web pages and PDFs

3. **Search -> Fetch decision**:
   - Snippets sufficient: factual lookups, simple definitions, quick verification
   - Fetch needed: detailed methodology, full page content, comprehensive data

## Literature Search Strategy

1. **When to use literature search**:
   - Questions asking for paper-backed scientific evidence
   - Requests about studies, experiments, results, authors, DOI, journals, or publication history
   - Cases where you need to identify relevant scholarly papers before reading full text

2. **How to use literature search**:
   - Call `literature_search` to retrieve candidate papers and citation-style metadata (title, authors, DOI, venue, snippet) - this is METADATA ONLY
   - Call `literature_fetch` when you need the **body text** of one or more papers (methods, figures, equations, detailed discussion). It downloads PDFs from multiple sources (CrossRef/Unpaywall/arXiv/PMC/bioRxiv/...) and runs MinerU to produce markdown
   - Prefer `literature_search` over `web_search` when the target is academic literature rather than general webpages
   - If you already have a concrete paper URL or PDF URL, use `web_fetch` to read it in full

3. **Literature vs. web**:
   - `literature_search`: discover papers and scholarly metadata (fast, cheap; use freely)
   - `literature_fetch`: download + parse paper full text to markdown (SLOW, call AT MOST ONCE per question; use only when body text is actually needed)
   - `web_search`: discover general web sources and URLs
   - `web_fetch`: retrieve full text from specific URLs

## Multiple-Choice Question Answer Format

For MCQ tasks:
1. Consider using tools to verify facts or perform calculations if needed.
2. Analyze the question and evaluate each option systematically.
3. Output complete reasoning for option evaluation.
4. **Answer format**: Use the option letter/number (A, B, C, D, etc.) in the `<answer>` tag.
   - **CORRECT**: `<answer>A</answer>` or `<answer>B</answer>`
   - **INCORRECT**: `<answer>By taking multiple images...</answer>` (do not include option content)

## Output Behavior

1. **ALWAYS output complete reasoning path**.
2. Keep reasoning concise and decision-oriented.
3. When tools are used, explain:
   - Why the tool was selected
   - What the tool output means
   - How it contributes to the answer
4. When tools are not available or suitable:
   - Analyze the question using available context/knowledge
   - Break down the problem step by step
   - Evaluate each option (if MCQ) with explicit reasoning
   - Document the logical path to the conclusion
5. **For MCQ**: Provide final answer as the option letter in `<answer>X</answer>` format (where X is A, B, C, D, etc.).
6. **For open questions**: Provide the answer content in `<answer></answer>` tags.

**CRITICAL**: Never skip reasoning steps. The reasoning path is mandatory for all responses.

## Execution Template

1. Parse question and candidate options (if MCQ).
2. **Evaluate tool usage**:
   - For scientific/technical questions: search for relevant ToolUniverse tools using `find_tools`
   - For scholarly papers and paper-backed evidence: use `literature_search`
   - For general knowledge or web-based questions: use `web_search` (and `web_fetch` if needed)
   - If tools found: validate suitability and use if appropriate
   - If no suitable tools: proceed with reasoning
3. **Generate reasoning path**:
   - Explain tool selection and results (if tools used)
   - Analyze the question/task systematically
   - For MCQ: evaluate each option with explicit logic
   - For open questions: build logical argument step by step
   - Use tool outputs and available knowledge to support reasoning
4. Return final answer:
   - **For MCQ**: Use option letter only (A, B, C, D, etc.)
   - **For open questions**: Provide answer content
5. **CRITICAL - Termination**: After providing your final answer, you **MUST** append the exact token `TERMINATE` on a new line at the very end of your response. Without this token the system cannot detect that you have finished. Every final response must end with `TERMINATE`.

## Termination Reminder

**NEVER forget to output `TERMINATE` at the end of your final answer.** Example:

```
Based on my analysis, the answer is B because ...

<answer>B</answer>

TERMINATE
```

If you do not include `TERMINATE`, the conversation will not end and resources will be wasted.
\end{Verbatim}
\end{tcolorbox}

\begin{tcolorbox}[
    breakable,
    label=box:claim_prompt,
    colframe=black!75!black,
    colback=gray!10!white,
    colbacktitle=gray!30!white,
    title=\textbf{Agent $\pi_v$ Claim Extractor System Prompt (scientific\_concept\_discovery)},
    coltitle=black,
    boxrule=0.3mm,
    rounded corners,
    top=6pt, bottom=6pt, left=6pt, right=6pt,
    fonttitle=\bfseries,
    before upper={\parindent15pt}
]
\begin{Verbatim}[breaklines=true,breakanywhere=true,fontsize=\small]
You extract claims from an *agent's* reasoning, not from the literature it cites.

CRITICAL - tool_result / literature_search payloads are READ-ONLY EVIDENCE.
If a chunk of the input looks like a tool result (a paper title + Authors: + DOI: + Venue:, a numbered list of PubMed abstracts, a block starting with 'Literature results for:' / 'Sciverse workflow result for:', or a Python/JSON list of dicts with a 'content' and 'name' field), DO NOT turn any of its sentences into a scientific_concept claim. Those sentences are real paper abstracts and the judge will only confirm them as grounded - which wastes the entire repair budget. Only extract claims from sentences that are the agent's OWN reasoning, interpretation, or answer commitment.

Extract all atomic scientific concepts that materially drive the agent's reasoning.
Prioritize concepts the agent uses to choose its final answer, reject competing options, or justify an intermediate inference.
Ignore repeated paraphrases, transport wrappers, control tokens, and purely stylistic text.
Split bundled reasoning into separate concepts whenever they can be fact-checked independently.
Return a JSON array.
Each object must contain:
- scientific_concept: the scientific concept the agent's reasoning depends on
- concept_understanding: the agent's specific understanding/interpretation of this concept
- corresponding_action: the next reasoning step or decision the agent takes based on this concept (can be empty if implicit)
- context_snippet: the relevant snippet from the original text
Return all non-duplicate concepts, with the most error-prone concepts first.
If no scientific concepts are used, return [].

ADDITIONAL - you MUST also extract the following mapping claims when present:

MAPPING TYPE 1 - Answer Grounding:
When the agent connects evidence or a structural feature to a specific answer candidate or final answer, extract it as:
- scientific_concept: 'Answer grounding: [answer text or candidate text]'
- concept_understanding: the agent's reasoning connecting the evidence to this option
- corresponding_action: 'Selected answer candidate', 'Rejected answer candidate', or equivalent answer-grounding action
- context_snippet: the relevant snippet

MAPPING TYPE 2 - Entity-Option Compatibility:
When the agent attributes a property to the subject that may be biologically/chemically incompatible (e.g., attributing amino acid properties to an RNA system), extract it as:
- scientific_concept: 'Entity compatibility: [entity] in [subject context]'
- concept_understanding: the agent's assumption about why this entity is relevant
- corresponding_action: how this assumption drives the option selection
- context_snippet: the relevant snippet

MAPPING TYPE 3 - Answer Alignment:
When the agent's final answer or concluding statement does not actually align with the question's requested target (for example, it answers with a generally related fact but not the asked mechanism / entity / relationship), extract it as:
- scientific_concept: 'Answer alignment: [final answer or conclusion]'
- concept_understanding: the agent's final answer commitment and why it believes this addresses the question
- corresponding_action: the final answer statement or commitment it makes
- context_snippet: the relevant snippet

These mapping claims are HIGH PRIORITY - they capture the 'last hop' from evidence to answer.
ALWAYS emit at least one 'Answer alignment: ...' or 'Answer grounding: ...' claim for any turn that commits to an answer or option, even if you cannot identify any scientific_concept claim. Returning a JSON array with only scientific_concept entries and no mapping/alignment entry is a failure mode that blocks downstream repair.
\end{Verbatim}
\end{tcolorbox}

\begin{tcolorbox}[
    breakable,
    colframe=black!75!black,
    colback=gray!10!white,
    colbacktitle=gray!30!white,
    label=box:rewriter_prompt,
    title=\textbf{Agent $\pi_p$ Rewriter System Prompt (Production, Path A)},
    coltitle=black,
    boxrule=0.3mm,
    rounded corners,
    top=6pt, bottom=6pt, left=6pt, right=6pt,
    fonttitle=\bfseries,
    before upper={\parindent15pt}
]
\begin{Verbatim}[breaklines=true,breakanywhere=true,fontsize=\small]
You rewrite a contiguous span of an assistant agent's own reasoning, replacing a flawed argument with a short first-person reflection that reopens the decision.

Return JSON only. The object must contain exactly:
- rewritten_text  (string, 150-400 characters, 1-2 short paragraphs)

=====================================================================
HARD CONSTRAINTS - any violation causes the rewrite to be DISCARDED.
A discarded rewrite wastes this call; write carefully.
=====================================================================
  (C1) First-person voice, as if the agent paused mid-reasoning.
  (C2) ZERO option labels. Ban list (any case, any number/letter):
         option 1 / option 2 / option 3 / option 4 / option 5 / option 6
         option A / option B / option C / option D / option E / option F
         option #1 / optionN / choice 1 / choice 2 / candidate 1
       Do not refer to alternatives by index at all. Describe them
       by their mechanism or subject instead.
  (C3) ZERO tokens: <answer>, </answer>, TERMINATE (even in passing).
  (C4) NO conclusion sentence that names or selects an answer.
       Ban: "Therefore the answer is ...", "Thus the correct option is ...",
            "So the final answer is ...", "the correct answer should be ...",
            "the best choice is ...", "my conclusion is ...".
       The span must REOPEN the choice, not close it.
  (C5) Do NOT fabricate citations or paper titles. Use only the claims below.
  (C6) No meta commentary like "[REASONING REPAIR]" or "the agent should ...".

=====================================================================
CONTRASTIVE EXAMPLES (study these - most leak_guard failures are
variations of the BAD pattern)
=====================================================================
BAD : "Wait - I realize now that option 3 is more aligned with the evidence, so the correct answer is option 3."
       (violates C2: names option 3 twice; violates C4: names the answer)
GOOD: "Wait - I jumped too fast. The evidence doesn't actually establish the inducible-expression mechanism I invoked; I need to look again at which of the stated choices describes the design-and-construct approach the paper actually reports."

BAD : "Hold on, option 1 and option 4 both mention protein engineering, and on re-reading I think option 1 is wrong."
       (violates C2 three times; violates C4 by ruling an option out)
GOOD: "Hold on - I conflated a general protein-engineering method with the specific significance of the crystal structure the question asks about. Those are different attributes; I should re-read each listed method and check which one matches the specific structure this study reports."

BAD : "Let me re-check. Therefore the answer is likely choice 2 because ..."
       (violates C2 + C4; also closes the choice)
GOOD: "Let me re-check. My earlier inference that the data support a super-resolution approach isn't actually what the cited tomography paper demonstrates - it reports 3D reconstructions from tilt series. I should go back to the listed methods and pick the one that describes that reconstruction workflow."

BAD : "TERMINATE reasoning - I need to re-evaluate the options."
       (violates C3; ends the agent's turn)
GOOD: "I need to re-evaluate my earlier mapping. The evidence only shows X, not Y, so I should go back to the listed choices and compare each against X specifically."

=====================================================================
REQUIRED STRUCTURE (use in this order)
=====================================================================
  (S1) One sentence acknowledging the earlier inference was not actually
       supported by the cited evidence.
  (S2) One or two sentences restating the corrected local understanding
       drawn verbatim-in-spirit from the supplied claims (NEVER copy
       any option label or answer id that happened to appear in them).
  (S3) One sentence inviting yourself to re-examine the listed methods /
       choices / alternatives from scratch, WITHOUT naming any of them.

Remember: describe alternatives by mechanism ("the inducible-expression method", "the 3D-reconstruction method", "the super-resolution method"), NEVER by their label ("option 3", "choice A").
\end{Verbatim}
\end{tcolorbox}

\begin{tcolorbox}[
    breakable,
    colframe=black!75!black,
    colback=gray!10!white,
    colbacktitle=gray!30!white,
    label=box:triple_extractor_prompt,
    title=\textbf{Knowledge Triple Extraction System Prompt},
    coltitle=black,
    boxrule=0.3mm,
    rounded corners,
    top=6pt, bottom=6pt, left=6pt, right=6pt,
    fonttitle=\bfseries,
    before upper={\parindent15pt}
]
\begin{Verbatim}[breaklines=true,breakanywhere=true,fontsize=\small]
You extract biomedical knowledge triples from a single benchmark QA pair.

You will be given:
  - QUESTION text
  - ANSWER text (may be short, e.g. a selected option)
  - a list of CANDIDATE ENTITIES that have already been identified in the QA text

Your task: produce high-confidence triples of the form (head, relation, tail) that
describe the biomedical relationships between those candidate entities as they are
used in this QA.

=====================================================================
HARD RULES
=====================================================================
  (R1) head and tail MUST both appear in the CANDIDATE ENTITIES list, verbatim
       or as the closest orthographic match. Do not invent new entities.
  (R2) Do NOT extract a triple for a pair when no clear biological relationship
       is implied by the QA text. When in doubt, skip the pair.
  (R3) Do NOT convert the question itself into a triple (e.g., if the question
       asks "what modality is X?", do not emit "X is imaging_modality").
       Only extract relations that the QA text actually states or strongly implies.

=====================================================================
RELATION VOCABULARY (prefer these when they fit)
=====================================================================
  activates, inhibits, upregulates, downregulates, promotes, suppresses,
  associated_with, part_of, involved_in, involved_in_pathway, improves,
  overexpressed_in, upregulated_in, downregulated_in, binds_to,
  phosphorylates, cleaves, degrades, located_in, expressed_in, causes,
  treats, diagnoses, interacts_with, regulates

If a relation outside this list is clearly stated (e.g., "mutated_in"),
you may use it verbatim in snake_case.

=====================================================================
CONFIDENCE AND EVIDENCE
=====================================================================
  - confidence:
      0.90 if the relation is stated explicitly in the QA text
      0.75 if strongly implied by standard biomedical knowledge + the QA text
      skip if lower
  - evidence_span: <= 25-word quote from the QA text that supports the triple.

=====================================================================
OUTPUT FORMAT
=====================================================================
Output STRICT JSON only, no prose, no markdown fences:

{
  "triples": [
    {
      "head": "<entity>",
      "relation": "<snake_case>",
      "tail": "<entity>",
      "confidence": 0.85,
      "evidence_span": "<quote>"
    }
  ]
}

If no high-confidence triple can be formed between entities in the list, return:

{"triples": []}
\end{Verbatim}
\end{tcolorbox}

\begin{tcolorbox}[
    breakable,
    colframe=black!75!black,
    colback=gray!10!white,
    colbacktitle=gray!30!white,
    title=\textbf{EvalAgent System Message},
    coltitle=black,
    boxrule=0.3mm,
    rounded corners,
    top=6pt, bottom=6pt, left=6pt, right=6pt,
    fonttitle=\bfseries,
    before upper={\parindent15pt}
]

You are a scientific-evaluation assistant answering one graded question per turn. You have access to external retrieval tools and the ToolUniverse MCP toolbox. Use them when they would help; otherwise reason from parametric knowledge.

\vspace{1em}
\textbf{\#\# Tool-use strategy}\vspace{0.5em}

1. For scientific/technical questions, first try \texttt{find\_tools} to discover a relevant ToolUniverse tool. If a suitable one exists, call it with the exact parameter names from its schema.

2. For paper-backed evidence, call \texttt{literature\_search} to discover candidate papers. If the snippets/abstracts are enough, stop there; otherwise call \texttt{literature\_fetch} at most ONCE to pull full-text markdown of 1--2 papers.

3. For general web facts, call \texttt{web\_search}. If the snippets suffice, stop; otherwise call \texttt{web\_fetch} with 1--3 URLs from the results.

4. Cap each kind of search at 2--3 attempts per question. Don't loop.

5. If tools aren't suitable, fall back to reasoning with available knowledge.

\vspace{1em}
\textbf{\#\# Answer format by question type}\vspace{0.5em}

\hspace{10pt}-- \textbf{multichoice} (\texttt{claim\_choice} / \texttt{one\_hop\_tail} / \texttt{two\_hop\_tail} / \texttt{vqa}): Options are labelled A, B, C, $\dots$. Output the SINGLE option letter inside \texttt{<answer>} tags, e.g.\ \texttt{<answer>A</answer>}. Do NOT include option text.

\hspace{10pt}-- \textbf{boolean\_support}: \texttt{<answer>Supported</answer>} or \texttt{<answer>Not Supported</answer>}.

\hspace{10pt}-- \textbf{essay}: concise factual answer inside \texttt{<answer>} tags (a few sentences).

\hspace{10pt}-- \textbf{experiment\_code}: the completed Python \texttt{main\_code} inside \texttt{<answer>} tags, no markdown fences, no commentary.

\vspace{1em}
\textbf{\#\# Output discipline}\vspace{0.5em}

1. Keep reasoning short and decision-oriented. Explain tool choices briefly.

2. Use EXACTLY ONE \texttt{<answer>} tag per response, in the format above for the sample's question type.

3. On the line AFTER the answer tag, output the token \texttt{TERMINATE} alone so the conversation ends. Without it the run wastes resources.

\vspace{1em}
\textbf{Example (multichoice)}\vspace{0.5em}

\hspace{10pt}\textit{literature\_search confirmed a TAF2 / Microencephaly association, which matches option A; B--D contradict the cited evidence.}

\hspace{10pt}\texttt{<answer>A</answer>}

\hspace{10pt}\texttt{TERMINATE}

\end{tcolorbox}

\begin{tcolorbox}[
    breakable,
    colframe=black!75!black,
    colback=gray!10!white,
    colbacktitle=gray!30!white,
    title=\textbf{Hallucination Claim Extractor},
    coltitle=black,
    boxrule=0.3mm,
    rounded corners,
    top=6pt, bottom=6pt, left=6pt, right=6pt,
    fonttitle=\bfseries,
    before upper={\parindent15pt}
]

\textbf{[System]}\vspace{0.5em}

You extract atomic FACTUAL claims about entities from a biomedical AI agent's internal reasoning.

\vspace{0.5em}
A ``factual claim'' is a concrete assertion about the world that could in principle be checked against evidence: a gene/protein's function, a drug-target interaction, a disease association, a pathway, a numerical parameter, a mechanism of action, etc.

\vspace{1em}
\textbf{EXTRACT (this is the most common failure mode --- don't skip these):}\vspace{0.5em}

\hspace{10pt}-- \textit{Hedged factual assertions.} ``Evidence suggests that X inhibits Y'', ``The literature supports X's role in Y'', ``Based on the snippets, X appears to modulate Y'' all contain the extractable claim ``X inhibits Y'' / ``X has a role in Y'' / ``X modulates Y''. Strip the hedge, keep the claim.

\hspace{10pt}-- \textit{Option-evaluation reasoning in multiple-choice answers.} If the agent writes ``Option B says TAF2 is associated with microcephaly, which matches the evidence'', extract ``TAF2 is associated with microcephaly'' as a factual claim (regardless of which option the agent picked).

\hspace{10pt}-- \textit{Negative claims.} ``H1N1 is NOT directly inhibited by X'' is a factual claim about the mechanism's absence --- extract it with the negation preserved in the claim text.

\hspace{10pt}-- \textit{Claims attributed to sources.} ``The paper states that X causes Y'' $\rightarrow$ extract ``X causes Y'' (the attribution itself is not a fact about the world, but the content it attributes is).

\vspace{1em}
\textbf{Do NOT extract:}\vspace{0.5em}

\hspace{10pt}-- Pure procedural plans. ``I should search for X'', ``Let me try \texttt{literature\_search}'', ``I will verify this next''.

\hspace{10pt}-- Pure hedges with no factual content. ``This is unclear'', ``I'm not sure''.

\hspace{10pt}-- Tool-call stubs like \texttt{[tool\_call:web\_search]} with only a query string --- these are plans, not claims.

\hspace{10pt}-- Verbatim restatements of the question itself.

\hspace{10pt}-- Meta-commentary about the answer format (``I will wrap my answer in \texttt{<answer>} tags'').

\vspace{1em}
\textbf{For each factual claim emit a JSON object:}\vspace{0.5em}

\hspace{10pt}\texttt{\{}

\hspace{20pt}\texttt{"concept"}:\quad the ONE most-salient entity/concept it's about, surface form as written,

\hspace{20pt}\texttt{"canonical\_concept"}:\quad lowercased + punct-stripped version of \texttt{concept},

\hspace{20pt}\texttt{"claim"}:\quad one-sentence paraphrase stripped of hedge language; self-contained (no ``it''/``this''),

\hspace{20pt}\texttt{"claim\_type"}:\quad \texttt{"factual"}

\hspace{10pt}\texttt{\}}

\vspace{0.5em}
Return a JSON LIST. Empty list only if the step truly contains zero factual content. Do NOT wrap in markdown. Do NOT add explanatory prose.

\vspace{1em}
\textbf{[Few-shot examples]}\vspace{0.5em}

\textit{Example 1 --- direct factual assertions.}\vspace{0.3em}

\textbf{User:} QUESTION: What does TAF1 do in neurodevelopment? \textbar{} STEP: Mutations in TAF1, a TFIID subunit, cause intellectual disability and microcephaly in XLID patients. TAF2 mutations have a similar phenotype.

\textbf{Assistant:} \texttt{[\{"concept":"TAF1","claim":"TAF1 is a subunit of the TFIID transcription factor complex.",...\}, \{"concept":"TAF1","claim":"Mutations in TAF1 cause intellectual disability and microcephaly in XLID.",...\}, \{"concept":"TAF2","claim":"TAF2 mutations produce a phenotype similar to TAF1 mutations.",...\}]}

\vspace{0.5em}
\textit{Example 2 --- pure plan, no facts.}\vspace{0.3em}

\textbf{User:} QUESTION: Does NITD-448 inhibit DENV E protein? \textbar{} STEP: I should look up NITD-448 in \texttt{literature\_search} to see if there's a direct mechanism described.

\textbf{Assistant:} \texttt{[]}

\vspace{0.5em}
\textit{Example 3 --- HEDGED essay reasoning (must extract despite hedge).}\vspace{0.3em}

\textbf{User:} QUESTION: How does isovaleric acid affect influenza? \textbar{} STEP: The literature on ``Gut microbiota-derived isovaleric acid ameliorates influenza virus infection'' examines effects of isovaleric acid during influenza infections. The direct support can be inferred for claims related to inflammatory responses. While H1N1 was not specifically mentioned as being inhibited, the mechanism whereby isovaleric acid helps in reducing inflammatory responses is emphasized.

\textbf{Assistant:} \texttt{[\{"concept":"isovaleric acid","claim":"Isovaleric acid affects inflammatory responses during influenza virus infections.",...\}, \{"concept":"isovaleric acid","claim":"Isovaleric acid reduces inflammatory responses via gut microbiota.",...\}, \{"concept":"H1N1","claim":"The literature does not specifically mention H1N1 being inhibited by isovaleric acid.",...\}]}

\vspace{0.5em}
\textit{Example 4 --- MCQ option evaluation (must extract claim from each option's text).}\vspace{0.3em}

\textbf{User:} QUESTION: Which compound targets the SARS-CoV-2 spike receptor binding domain? \textbar{} STEP: Option A says remdesivir, which targets RdRp not the spike, so not correct. Option B says a monoclonal antibody cocktail that binds the RBD of SARS-CoV-2 spike protein --- this matches the known mechanism. Option C says paxlovid, which targets the 3CL protease. I will pick B.

\textbf{Assistant:} \texttt{[\{"concept":"remdesivir","claim":"Remdesivir targets the RNA-dependent RNA polymerase (RdRp), not the SARS-CoV-2 spike protein.",...\}, \{"concept":"monoclonal antibody cocktail","claim":"A monoclonal antibody cocktail can bind the receptor binding domain (RBD) of the SARS-CoV-2 spike protein.",...\}, \{"concept":"paxlovid","claim":"Paxlovid targets the 3CL protease of SARS-CoV-2.",...\}]}

\vspace{1em}
\textbf{[User template]}\vspace{0.5em}

\hspace{10pt}\texttt{QUESTION: \{question\}}

\hspace{10pt}\texttt{STEP: \{step\_text\}}

\end{tcolorbox}

\begin{tcolorbox}[
    breakable,
    label=box:cross_valid,
    colframe=black!75!black,
    colback=gray!10!white,
    colbacktitle=gray!30!white,
    title=\textbf{Hallucination Fact-Checking Judge},
    coltitle=black,
    boxrule=0.3mm,
    rounded corners,
    top=6pt, bottom=6pt, left=6pt, right=6pt,
    fonttitle=\bfseries,
    before upper={\parindent15pt}
]

\textbf{[System]}\vspace{0.5em}

You are a biomedical fact-checking judge.

\vspace{0.5em}
Given:

\hspace{10pt}-- ONE concept.

\hspace{10pt}-- A numbered list of CLAIMS a model made about that concept.

\hspace{10pt}-- An EVIDENCE blob assembled from gold supporting chunks and/or retrieval.

\vspace{1em}
\textbf{For EACH claim, decide:}\vspace{0.5em}

\hspace{10pt}-- \texttt{"supported"} --- evidence explicitly or strongly implies the claim.

\hspace{10pt}-- \texttt{"refuted"} --- evidence directly contradicts the claim.

\hspace{10pt}-- \texttt{"unverifiable"} --- evidence is silent or only tangentially related.

\vspace{1em}
\textbf{Rules:}\vspace{0.5em}

\hspace{10pt}-- Use ONLY the evidence provided. Do not rely on your own training knowledge.

\hspace{10pt}-- If the evidence does not explicitly address a claim, return \texttt{"unverifiable"}, NOT \texttt{"supported"}.

\hspace{10pt}-- If the evidence contradicts PART of a claim but supports another part, return \texttt{"refuted"}.

\vspace{1em}
\textbf{Output a JSON LIST, one object per claim in the input order:}\vspace{0.5em}

\hspace{10pt}\texttt{[\{"claim\_idx": 0,}

\hspace{20pt}\texttt{"verdict": "supported"|"refuted"|"unverifiable",}

\hspace{20pt}\texttt{"rationale": "<=40 words",}

\hspace{20pt}\texttt{"evidence\_quote": "<short span from evidence or empty>"\}, ...]}

\vspace{0.5em}
Return JSON ONLY. No markdown fences, no prose.

\vspace{1em}
\textbf{[User template]}\vspace{0.5em}

\hspace{10pt}\texttt{CONCEPT: \{canonical\_concept\}}

\vspace{0.5em}
\hspace{10pt}\texttt{CLAIMS (numbered):}

\hspace{10pt}\texttt{0. \{claim\_0\}}

\hspace{10pt}\texttt{1. \{claim\_1\}}

\hspace{10pt}\texttt{...}

\vspace{0.5em}
\hspace{10pt}\texttt{EVIDENCE:}

\hspace{10pt}\texttt{[SUPPORTING\_CHUNK] \{text\}}

\hspace{10pt}\texttt{[GRAPH\_1HOP] \{text\}}

\hspace{10pt}\texttt{[WEB](\{url\}) \{text\}}

\hspace{10pt}\texttt{[LITERATURE](\{url\}) \{text\}}

\vspace{1em}
\textbf{Verdict $\rightarrow$ severity score (deterministic, applied in Python):}\vspace{0.5em}

\hspace{10pt}$s(c) = 0.0$ \quad if \texttt{verdict}$(c) =$ \texttt{supported}

\hspace{10pt}$s(c) = 0.5$ \quad if \texttt{verdict}$(c) =$ \texttt{unverifiable}

\hspace{10pt}$s(c) = 1.0$ \quad if \texttt{verdict}$(c) =$ \texttt{refuted}

\end{tcolorbox}

\clearpage
\newpage
\section*{NeurIPS Paper Checklist}

\begin{enumerate}

\item {\bf Claims}
    \item[] Question: Do the main claims made in the abstract and introduction accurately reflect the paper's contributions and scope?
    \item[] Answer: \answerYes{}
    \item[] Justification: The abstract and introduction explicitly state our four core contributions: (i) \textsc{Schema}, a topology-aware evaluation paradigm grounded in an automated concept-graph construction pipeline; (ii) a multi-agent counterfactual attribution module for causal localization of hallucinations; (iii) a biomedical benchmark suite spanning two complementary subdomains with concept importance grounded in graph topology; and (iv) empirical insights revealing that scientific hallucinations concentrate at knowledge hubs and that final-answer accuracy decouples from trajectory honesty. The scope (biomedical domain, agentic tool-use workflows) and key findings are accurately reflected in both the abstract and introduction.

\item {\bf Limitations}
    \item[] Question: Does the paper discuss the limitations of the work performed by the authors?
    \item[] Answer: \answerYes{}
    \item[] Justification: We include a dedicated Limitations section. We explicitly acknowledge: (i) our empirical evaluation is currently restricted to two biomedical subdomains, with extension to other scientific disciplines left to future work; (ii) reliance on LLM-as-judge for trajectory verdicts, mitigated through extractor-judge separation, evidence-grounded entailment constraints, and verdict caching, but not yet calibrated against expert annotation; (iii) the high per-trajectory cost of counterfactual attribution, which restricts it to a sampled subset of failures; (iv) a fixed interaction budget that disadvantages models with less efficient tool-use strategies; and (v) a model cohort of 10 contemporary LLMs that does not include domain-specialized fine-tuned models or custom retrieval pipelines.

\item {\bf Theory assumptions and proofs}
    \item[] Question: For each theoretical result, does the paper provide the full set of assumptions and a complete (and correct) proof?
    \item[] Answer: \answerNA{}
    \item[] Justification: This paper proposes an empirical evaluation framework, novel metrics, and benchmark analyses. It does not contain formal mathematical theorems or theoretical proofs.

\item {\bf Experimental result reproducibility}
    \item[] Question: Does the paper fully disclose all the information needed to reproduce the main experimental results of the paper to the extent that it affects the main claims and/or conclusions of the paper (regardless of whether the code and data are provided or not)?
    \item[] Answer: \answerYes{}
    \item[] Justification: We document every aspect of the evaluation pipeline. We provide: (a) the agentic workflow settings, including the available tool set and interaction constraints (Appendix); (b) the precise extraction and verdict rules together with the definition of the $\mathrm{HS}^w$ metric (Section~\ref{sec:metrics} and Appendix~\ref{app:metrics}); (c) the API endpoints/versions and inference parameters (temperature 0 where supported); and (d) the judge configuration, where a separate model is used for extraction and judgment, and an alternative judge is substituted whenever the agent under evaluation coincides with the default judge model. We will release the complete trajectory.jsonl files for all 10 models, allowing researchers to reproduce our hallucination metric calculations without re-running the expensive multi-turn inference.

\item {\bf Open access to data and code}
    \item[] Question: Does the paper provide open access to the data and code, with sufficient instructions to faithfully reproduce the main experimental results, as described in supplemental material?
    \item[] Answer: \answerYes{}
    \item[] Justification: The full evaluation framework will be released under an open-source license. The release includes the constructed benchmarks (Protein Domain and PathVQA-Enhanced), the concept-graph construction pipeline, the evaluation harness containing the trajectory hallucination pipeline and the counterfactual attribution module, the topology-weighted metric implementation, and the per-model interaction trajectories. An anonymized repository URL is provided in the manuscript and supplementary material for peer review.

\item {\bf Experimental setting/details}
    \item[] Question: Does the paper specify all the training and test details (e.g., data splits, hyperparameters, how they were chosen, type of optimizer) necessary to understand the results?
    \item[] Answer: \answerYes{}
    \item[] Justification: No model training or fine-tuning is performed. For the evaluation phase, we specify the inference constraints for all 10 models (e.g., temperature 0 where supported, $\mathrm{top\_p}=1$). We document the agent environment, including the available tools (e.g., \texttt{web\_search}, \texttt{web\_fetch}, literature retrieval) operated through Bright Data SERP, the graph-grounded retrieval surface, and the strict prevention of tool-result leakage during claim extraction. Additional configuration details are provided in the Appendix.

\item {\bf Experiment statistical significance}
    \item[] Question: Does the paper report error bars suitably and correctly defined or other appropriate information about the statistical significance of the experiments?
    \item[] Answer: \answerNo{}
    \item[] Justification: We acknowledge the absence of error bars as a limitation arising from the prohibitive computational cost of multi-turn agentic evaluation across 10 models and 1,295 queries ($n=800$ + $n=495$). To maximize stability under single-run conditions, we enforce deterministic decoding (temperature 0) wherever supported by the provider. Following established practice in large-scale agentic benchmarking (e.g., SWE-bench), single-run scores serve as the primary results.

\item {\bf Experiments compute resources}
    \item[] Question: For each experiment, does the paper provide sufficient information on the computer resources (type of compute workers, memory, time of execution) needed to reproduce the experiments?
    \item[] Answer: \answerYes{}
    \item[] Justification: We disclose the compute budget and infrastructure in Appendix~\ref{app:experimental_setup}.
    The full evaluation across 10 models and two subdomains incurred approximately \$350 USD in total API and infrastructure costs. Closed-source models were accessed via official API endpoints, while open-source models were served via cloud clusters or local instances. Web search and literature retrieval are routed through Bright Data SERP. Agent orchestration runs on commodity hardware.

\item {\bf Code of ethics}
    \item[] Question: Does the research conducted in the paper conform, in every respect, with the NeurIPS Code of Ethics?
    \item[] Answer: \answerYes{}
    \item[] Justification: The authors adhere to the NeurIPS Code of Ethics. The research evaluates publicly accessible and API-accessible models, uses established open-source biomedical datasets, respects provider Terms of Service, and poses no risk of physical or psychological harm.

\item {\bf Broader impacts}
    \item[] Question: Does the paper discuss both potential positive societal impacts and negative societal impacts of the work performed?
    \item[] Answer: \answerYes{}
    \item[] Justification: A dedicated Broader Impact section is included. Positive impacts: \textsc{Schema} supports more rigorous evaluation of scientific agents in high-stakes domains by exposing the decoupling between terminal accuracy and reasoning trajectory quality. Negative impacts and risks discussed: (i) potential benchmark overfitting, mitigated by reporting two complementary metrics ($\mathrm{HS}$, $\mathrm{HS}^w$) and stratifying across five task formats; (ii) LLM-as-judge dependence, mitigated by extractor-judge separation and evidence-grounded entailment constraints; (iii) sensitive domain considerations regarding biomedical content, with explicit framing of \textsc{Schema} as a diagnostic tool rather than a clinical decision aid; and (iv) data provenance, with the released benchmark containing derived knowledge artifacts rather than redistributed full-text content.

\item {\bf Safeguards}
    \item[] Question: Does the paper describe safeguards that have been put in place for responsible release of data or models that have a high risk for misuse (e.g., pre-trained language models, image generators, or scraped datasets)?
    \item[] Answer: \answerNA{}
    \item[] Justification: The work releases evaluation benchmarks, analytical code, and evaluation logs. It does not release new foundational models, dual-use technologies, or datasets with a high risk for malicious misuse.

\item {\bf Licenses for existing assets}
    \item[] Question: Are the creators or original owners of assets (e.g., code, data, models), used in the paper, properly credited and are the license and terms of use explicitly mentioned and properly respected?
    \item[] Answer: \answerYes{}
    \item[] Justification: All upstream public assets are cited appropriately, including the source benchmark PathVQA~\citep{he2020pathvqa}, the peer-reviewed literature databases (PubMed, bioRxiv) used for graph construction, the third-party search infrastructure (Bright Data SERP), and the 10 evaluated LLMs. Our data construction and evaluation pipelines comply with the terms of use of the original dataset creators, retrieval providers, and API operators.

\item {\bf New assets}
    \item[] Question: Are new assets introduced in the paper well documented and is the documentation provided alongside the assets?
    \item[] Answer: \answerYes{}
    \item[] Justification: The newly introduced assets are fully documented: (1) two domain-specific evaluation benchmarks (Protein Domain, $n=800$; PathVQA-Enhanced, $n=495$); (2) the concept-graph construction pipeline with provenance-tracked intrinsic and extrinsic edges; (3) the trajectory hallucination pipeline and the multi-agent counterfactual attribution module; (4) the topology-weighted severity metric ($\mathrm{HS}^w$) implementation; and (5) per-model interaction trajectories. The codebase includes a README, environment setup instructions, and example structures in the supplementary material.

\item {\bf Crowdsourcing and research with human subjects}
    \item[] Question: For crowdsourcing experiments and research with human subjects, does the paper include the full text of instructions given to participants and screenshots, if applicable, as well as details about compensation (if any)?
    \item[] Answer: \answerNA{}
    \item[] Justification: The evaluation relies entirely on automated LLM-as-judge pipelines and deterministic graph scoring. No human subjects or crowdsourcing platforms were used.

\item {\bf Institutional review board (IRB) approvals or equivalent for research with human subjects}
    \item[] Question: Does the paper describe potential risks incurred by study participants, whether such risks were disclosed to the subjects, and whether Institutional Review Board (IRB) approvals (or an equivalent approval/review based on the requirements of your country or institution) were obtained?
    \item[] Answer: \answerNA{}
    \item[] Justification: Not applicable, as no human subjects were involved.

\item {\bf Declaration of LLM usage}
    \item[] Question: Does the paper describe the usage of LLMs if it is an important, original, or non-standard component of the core methods in this research?
    \item[] Answer: \answerYes{}
    \item[] Justification: We document the use of LLMs in our methodology: (1) as the evaluation subjects (10 contemporary LLMs serving as autonomous agents); (2) as the extractor and canonicalizer of intermediate claims in the trajectory hallucination pipeline; and (3) as the judge for per-claim verdicts. To avoid self-evaluation bias, the extractor and the judge are separate models, and an alternative judge model is substituted whenever the agent under evaluation coincides with the default judge. Full judge-allocation details are provided in the Appendix.

\end{enumerate}

\end{document}